%% file: Tuning_paper_v5_ML.TeX
\documentclass[twoside,11pt]{article}

\usepackage{natbib}
\pdfoutput=1
\usepackage[english]{babel}
\usepackage{amsmath}
\usepackage{amssymb}%
\usepackage{mathrsfs}
\usepackage{amsbsy}
\usepackage{amsthm}
\usepackage{geometry}
\usepackage{hyperref}

\usepackage{float}

\usepackage{enumerate}
\usepackage{enumitem}
\usepackage{booktabs}
\usepackage{multirow}

\usepackage{relsize}
\usepackage{graphics}

\usepackage[]{graphicx}

\usepackage{accents} 

\usepackage[table]{xcolor}

\usepackage{url}
\usepackage{array}
\usepackage{bookmark}

\usepackage{caption}
\usepackage{subcaption}

\usepackage[modulo]{lineno}

\newcommand\bsln{1.5} 

\renewcommand{\baselinestretch}{\bsln}

\newcommand{\blind}{0}

\title{Blocked Cross--Validation: A Precise and Efficient Method for Hyperparameter Tuning}
\if0\blind
\author{Giovanni Maria Merola  \\
       MIT - World Peace University, Pune, India\\
       ORCID ID  0000-0003-2539-9225\\
       \href{merolagio@gmail.com}{\nolinkurl{merolagio@gmail.com}}
       }
\fi

\date{\vspace{-5ex}}
\providecommand{\keywords}[1]{\textbf{\textit{Keywords --- }} #1}

\input{Tuning_macros.TeX}

\begin{document}

\maketitle

\begin{abstract}
Hyperparameter tuning plays a crucial role in optimizing the performance of predictive learners. Cross--validation (CV) is a widely adopted technique for estimating the error of different hyperparameter settings. Repeated cross--validation (RCV) has been commonly employed to reduce the variability of CV errors. In this paper, we introduce a novel approach called blocked cross--validation (BCV), where the repetitions are blocked with respect to both CV partition and the random behavior of the learner. Theoretical analysis and empirical experiments demonstrate that BCV provides more precise error estimates compared to RCV, even with a significantly reduced number of runs. We present extensive examples using real--world data sets to showcase the effectiveness and efficiency of BCV in hyperparameter tuning. Our results indicate that BCV outperforms RCV in hyperparameter tuning, achieving greater precision with fewer computations.
\end{abstract}

\section{Introduction}
Cartesian grid search using K--fold cross--validation \citep[CV,][]{sto74, all} is a simple widely used method for fine-tuning predictive learners,\footnote{We refer to algorithms that use a set of features to predict a response generically as ``predictive learners'' or simply ``learners''. We reserve the term ``model'' for a theoretical description of the relationship among variables.} with hyperparameters $\bth = (\bth_1, \bth_2\ldots)$.
CV prediction error (\ecv) is computed for all combinations of preselected values of the learner's hyperparameters, called \textit{settings}, and the setting that corresponds to the lowest error is chosen as the most promising. This discrete search is unlikely to find the best setting, but it is computationally feasible and it is expected to find good settings.

At the moment it is still unclear what \ecv estimates. According to \cite{esl} and \cite{bat} it estimates the unconditional prediction error rather than the actual prediction error conditional on the training data. In this paper we take \ecv at face value, as a measure of a learner's predictive accuracy, irrespective of what it estimates. Thus, we focus on the properties of \ecv within the available training data, rather than those concerning all possible training and future data. For this reason, we omit the reference to the training set $T$ in our notation.

Resampling methods, including CV, introduce both bias and variance in prediction error estimates \cite[][among others]{mol, esl, arl, kuh}. CV is considered a compromise between the highly biased leave--one--out CV and the highly variable hold--out validation set estimates \citep{esl}. \ecv's bias is caused by the deterioration of the learner's performance when trained on fewer data than are available. The magnitude of this bias cannot be easily estimated because it depends on a variety of variables, including the number of cases, type of learner, structure of the data and number of folds \citep[][and references therein]{arl}.
Since \ecv's bias concerns generalization error and is invariant to different settings for the same CV partitioning strategy \citep{arl}, it is irrelevant to hyperparameter tuning and we do not discuss it further.  By 'partitioning strategy' (or simply 'strategy'), we refer to the number of folds and the type of sampling (for instance, simple random or stratified sampling) employed to partition the training data. Here we assume that, when for tuning the hyperparameters of a learner, \ecv are computed using the same strategy.
\subsection{Variance of \ecv}
Several papers, as referenced and discussed in \citet{you21}, have investigated the variance of \ecv within the training data. These works attempt to estimate the sampling variance of \ecv, which is also affected by the same issues as the bias. For instance, \cite{ben} showed that there is no unbiased estimate of \ecv's variance. In this paper, we take a different approach, considering \ecv's variance as a nuisance and proposing a method for reducing it to obtain more precise estimates of hyperparameter's effects and prediction error.

We devised the Blocked Cross--Validation (BCV) method based on the consideration that CV is a deterministic procedure with only two sources of variability: the selection of CV partition and the potential random behavior exhibited by the learner.

The latter source of variability, which we loosely identify with the algorithm's initial random seed, arises in predictive learners that incorporate random elements, such as resampling, variable selection for individual trees or random drop-outs, for example.
CV's variability can result in inconsistent outcomes when selecting optimal settings,
because a setting that yields a lower \ecv may perform worse when evaluated using a different CV partition and initial seed value. Consequently, suboptimal settings can be selected during grid searches due to the use of \ecv s computed on varying CV partitions and initial seeds.

As the CV partitioning process is not part of a learner's behavior, it is crucial to eliminate the variability introduced by it when comparing \ecv s computed under different settings.
On the other hand, the random behavior is more akin to a random hyperparameter, because the learner used in production will be trained on a single initial seed. However, tuning the initial seed is pointless because its effect on the learner's performance is random by definition. While it is commonly believed that the choice of the starting seed has little impact on the learners' predictive performance, different initial seeds can actually determine different optimal settings for the same CV partition. Therefore, to compare the \ecv across different settings, it is important to eliminate, or at least reduce, the effect of both the CV partition and initial seed, which can alter the value of the \ecv.

\subsection{Experimental Designs}
In traditional statistics, a cartesian grid is commonly referred to as an experimental design and is typically analyzed as a linear model using Analysis of Variance (ANOVA). ANOVA breaks down the response variance into the sum of the variance due to the experimental factors and the variance due to the error. When known variables that are of no interest, referred to as \textit{nuisance variables}, increase the response variability, experimental designs can reduce their effect by estimating and removing the variance that they induce, either by
randomization or blocking. Randomization involves repeatedly measuring the response for each experimental setting, allowing the nuisance variables to take random values, and then averaging the results. Randomization can produce imbalances between treatments, and the sources of error are completely confounded with each other and cannot be isolated and removed. In blocking, on the other hand, the same experiments are repeated with the nuisance variables set to different fixed values. The blocking principle is to compare like with like. In fact, blocking enables the comparison of the responses of experiments to the same value of the nuisance variable. Furthermore, with blocking, the experimental variance between blocks can be isolated and removed. Blocking is far more effective than randomization at reducing the variance of the estimates, as noted by several highly respected sources  \citep{fis, cox, box, mon, wu}. ``Block what you can control, randomize what you cannot,'' says an old adage in Statistics.
\subsection{Blocked Cross-Validation}
In grid search tuning, CV partition and initial seed are nuisance variables that are responsible for all the variability of the \ecv.  Repeated CV (RCV) is a popular and straightforward randomization method used to reduce the variance of the \ecv in grid search.

In this paper we propose a new method called Blocked CV (BCV), in which a hyperparameter grid is blocked with respect to CV partitioning and algorithm initial seed. The blocking on the random behavior extends to the computation of the error in each fold, whereas, to our knowledge, the \ecv for given settings are usually obtained by averaging the errors of different folds computed using different initial seeds.
BCV reduces the standard error of the estimated \ecv by partially eliminating the variance due to the CV partition and initial seed, allowing for a more precise comparison of the \ecv of different settings.
BCV is also computationally more efficient than RCV because more precise estimates can be obtained with fewer runs and it requires performing fewer data partitions. To the best of our knowledge, no other work has considered using blocking to reduce the variability of the \ecv in learner tuning. Another advantage of BCV over other grid search methods is that it improves the precision of the hyperparameter effects estimates and of their significance tests. Classic F-tests are not appropriate for the analysis of \ecv experiments \citep{die}. However, permutation tests can be used to test the effects of hyperparameters.
\subsection{Manuscript Roadmap}
In the next section we will illustrate  how computing \ecv using different CV partitions and algorithm's initial seeds can lead to choosing a suboptimal settings in a grid-search, and how using BCV leads to computing more precise estimates of the ErrCV, hence more reliable results. We also show how BCV is more efficient than RCV for this task. For this purpose, we will use a small data set for tuning a Random Forest (RF).

Although BCV is a versatile methodology applicable to any predictive learner and data set of any size, we have chosen to showcase results focusing on tuning RF with small data sets. This choice was motivated by computational convenience and the recognition that small data sets are common in fields where measuring the response is challenging or time-consuming, such as medical, environmental, and neuroscience research. Unfortunately, these small data sets are sometimes overlooked in research. We chose to run the examples on tuning RFs only, because RF is a popular predictive learner that incorporates random behavior, and exploring tuning results for multiple learners would have been impractical within the scope of this paper.


In Section \ref{sec:relwork}, we provide an overview of the relevant literature. Section \ref{sec:theory} presents the theoretical justification for favoring BCV over RCV in parameter tuning, were we also give a statistical model for RCV, which we could not find in past literature. Moving on to Section \ref{sec:dataanal}, we report the results obtained from analyzing five additional data sets. Finally, in Section \ref{sec:remarks}, we offer concluding remarks and further considerations.
\section{Motivating example}\label{seq:motex}
The Sonar data set \citep{Sonar} consists of a small sample of 208 instances with 60 frequency readings reflected off objects that could be cylindrical rocks or mines. The objective is to classify the objects as rocks or mines based on the frequencies. For this classification task, the \ecv is given by the proportion of misclassified instances.
Table  \ref{tab:so_param} displays the values of the four hyperparameters used to construct the cartesian grid for tuning the RF. The settings with \emph{replace = F} and \emph{sample.fraction = 1} were removed, which left a total of 84 different settings.
Note that the values of the $mtry$ hyperparameter are relatively small compared to the total number of features due to their high correlation. All RFs were trained using a large value of 2000 trees. It is important to note that when comparing the same designs with different sampling strategies on the same data set, we use the same blocking seeds.
\renewcommand{\baselinestretch}{0.9}
{
\begin{table}[H]
  \centering
\scriptsize{
    \begin{tabular}{lll}
    \toprule
    Parameter & name  & values \\
    \midrule
    number of variables per tree & mtry  & 5 10 20 \\
    minimum node size & min.node.size & 3 5 10 15 \\
    sample with replacement & replace & T F \\
    sampling fraction & sample.fraction & 0.5 0.7 0.9 1.0 \\
    \bottomrule
    \end{tabular}%
    }%
      \caption{Values used for tuning the random forest.}
  \label{tab:so_param}%
\end{table}%
}
\renewcommand{\baselinestretch}{\bsln}

Throughout the paper, we will use the following notation to distinguish between different partitioning strategies. The number of folds will be represented as "k," while "SRS" or "STS" will indicate whether simple random or stratified subsampling was used, respectively. To differentiate designs, we will use "BCV YxZ" to represent a design blocked with Y CV seeds and Z RF seeds, and "RCV NRep" for RCV with N repetitions.
For instance, "5-BCV STS 4x2" means the tuning involves 5-fold CV with stratified sampling, blocking with four CV seeds and two RF seeds. Certain elements of this notation may be excluded if they're not required in the given context.

\begin{figure}[H]
  \centering
  \includegraphics[width=0.75\textwidth]{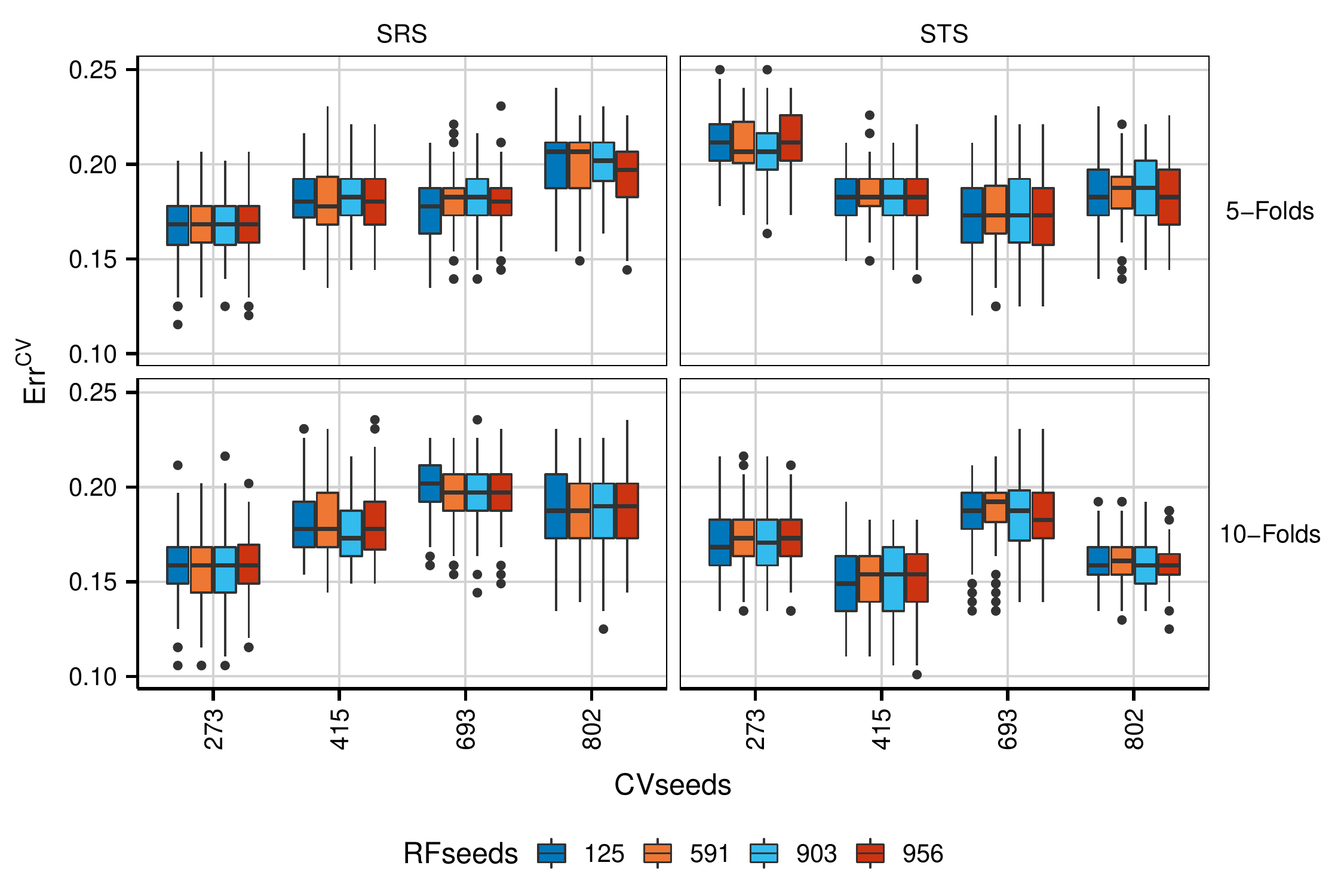}
  \caption{Sonar data: distribution of the \ecv computed with BCV 4 x 4 and different partitioning strategies.}\label{fig:so_tune_bxplAll_seeds}
\end{figure}

The box--and-whiskers plots in Figure \ref{fig:so_tune_bxplAll_seeds} show the distribution of the BCV 4x4 \ecv s, computed using different partitioning strategies
The erratic effect of the different CVseeds and, to a smaller scale, of the RFseeds on the \ecv clearly shows in this plot. Note that the differences span over 10\% misclassification error.

Table \ref{tab:sonar_best1} the impact of of using different seed combinations on the CV results. The settings shown displayed those that give the lowest \ecv when employing 5-BCV STS with different seed combinations, corresponding to the distributions depicted in the top-right panel of \ref{fig:so_tune_bxplAll_seeds}. This illustration serves as an example, as similarly conflicting outcomes can be observed with other partitioning strategies. Certain seed combinations yield two or three equally optimal settings, while in many instances, different seed combinations result in distinct best settings, with several of these not being either 61 or 64, which are the overall optima. All minima are associated with sampling without replacement, and the majority correspond to a 90\% sampling fraction; the number of variables attempted is always equal to five but for two seeds combinations, when it is equal to ten. The minimum node size display more variability within these minima.

\renewcommand{\baselinestretch}{0.5}
\begin{table}[H]
  \centering
{\scriptsize
    \begin{tabular}{rrrrrrrrl}
\cmidrule{1-8}    \multicolumn{1}{l}{mtry} & \multicolumn{1}{l}{min.node.size} & \multicolumn{1}{l}{replace} & \multicolumn{1}{l}{sample.fraction} & \multicolumn{1}{l}{CVseeds} & \multicolumn{1}{l}{RFseeds} & \multicolumn{1}{l}{SettingNo} & \multicolumn{1}{l}{\ecv} &  \\
\cmidrule{1-8}    5     & 3     & F & 0.7   & 273   & 125   & 37    & 17.8\% & * \\
    5     & 5     & F & 0.9   & 273   & 125   & 64    & 17.8\% & * \\
\cmidrule{1-8}    5     & 5     & F & 0.9   & 273   & 591   & 64    & 17.3\% &  \\
    5     & 3     & F & 0.9   & 273   & 903   & 61    & 16.3\% &  \\
\cmidrule{1-8}    5     & 3     & F & 0.9   & 273   & 956   & 61    & 17.3\% & * \\
    5     & 5     & F & 0.9   & 273   & 956   & 64    & 17.3\% & * \\
\cmidrule[0.85pt]{1-8}    5     & 3     & F & 0.9   & 415   & 125   & 61    & 14.9\% &  \\
\cmidrule{1-8}   5     & 3     & F & 0.9   & 415   & 591   & 61    & 14.9\% & * \\
    5     & 5     & F & 0.9   & 415   & 591   & 64    & 14.9\% & * \\
\cmidrule{1-8}    5     & 3     & F & 0.9   & 415   & 903   & 61    & 14.4\% &  \\
    5     & 5     & F & 0.9   & 415   & 956   & 64    & 13.9\% &  \\
\cmidrule[0.85pt]{1-8}    5     & 3     & F & 0.9   & 693   & 125   & 61    & 12.0\% & * \\
    5     & 5     & F & 0.9   & 693   & 125   & 64    & 12.0\% & * \\
\cmidrule{1-8}    5     & 3     & F & 0.9   & 693   & 591   & 61    & 12.5\% & * \\
    5     & 5     & F & 0.9   & 693   & 591   & 64    & 12.5\% & * \\
    5     & 10    & F & 0.9   & 693   & 591   & 67    & 12.5\% & * \\
\cmidrule{1-8}    5     & 5     & F & 0.9   & 693   & 903   & 64    & 12.5\% &  \\
\cmidrule{1-8}    5     & 5     & F & 0.7   & 693   & 956   & 40    & 12.5\% & * \\
    5     & 5     & F & 0.9   & 693   & 956   & 64    & 12.5\% & * \\
    5     & 10    & F & 0.9   & 693   & 956   & 67    & 12.5\% & * \\
\cmidrule[0.85pt]{1-8}    5     & 3     & F & 0.9   & 802   & 125   & 61    & 13.9\% & * \\
    5     & 5     & F & 0.9   & 802   & 125   & 64    & 13.9\% & * \\
\cmidrule{1-8}    5     & 5     & F & 0.9   & 802   & 591   & 64    & 13.9\% &  \\
\cmidrule{1-8}    10    & 3     & F & 0.9   & 802   & 903   & 62    & 14.4\% & * \\
    10    & 5     & F & 0.9   & 802   & 903   & 65    & 14.4\% & * \\
\cmidrule{1-8}    10    & 3     & F & 0.9   & 802   & 956   & 62    & 14.4\% &  \\
\cmidrule{1-8}
\end{tabular}%
}
  \caption{Sonar data: settings that yielded the lowest \ecv for different combinations of seeds using 5--BCV STS 4x4. Multiple minima in the same block are indicated with an asterisk.}
  \label{tab:sonar_best1}%
\end{table}%
\renewcommand{\baselinestretch}{\bsln}

Table \ref{tab:so_2best} presents the two settings that resulted in the lowest \ecv averaged over the seeds  combinations, $\ecvb$, for each partitioning strategy. Setting number 61 consistently outperforms the others as the best choice across all strategies, while setting number 64 emerges as the second-best option. The only distinction between these two settings is the minimum node size, with setting 61 having a value of three and setting 64 having a value of five. Notably, the 10-folds STS strategy exhibits the lowest $\ecvb$ values, while the 5-folds STS strategy yields higher $\ecvb$  compared to its corresponding SRS counterpart.
\renewcommand{\baselinestretch}{0.7}
\begin{table}[H]
  \centering
  {\scriptsize
    \begin{tabular}{rrrrrrr}
    \toprule
   \multicolumn{1}{l}{mtry} & \multicolumn{1}{l}{min.node.size} & \multicolumn{1}{l}{replace} & \multicolumn{1}{l}{sample.fraction} & \multicolumn{1}{l}{TreatNo} & \multicolumn{1}{l}{$\ecvb$} & \multicolumn{1}{l}{Means} \\
    \midrule
    5     & 3     & F & 0.9   & 61    & 0.1402 & \multirow{2}[2]{*}{Overall} \\
    5     & 5     & F & 0.9   & 64    & 0.1418 &  \\
    \midrule
    5     & 3     & F & 0.9   & 61    & 0.1424 & \multirow{2}[2]{*}{5--BCV SRS} \\
    5     & 5     & F & 0.9   & 64    & 0.1436 &  \\
    \midrule
    5     & 3     & F & 0.9   & 61    & 0.1481 & \multirow{2}[2]{*}{5--BCV STS} \\
    5     & 5     & F & 0.9   & 64    & 0.1481 &  \\
    \midrule
    5     & 3     & F & 0.9   & 61    & 0.1394 & \multirow{2}[2]{*}{10--BCV SRS} \\
    5     & 5     & F & 0.9   & 64    & 0.1430 &  \\
    \midrule
    5     & 3     & F & 0.9   & 61    & 0.1310 & \multirow{2}[2]{*}{10--BCV STS} \\
    5     & 5     & F & 0.9   & 64    & 0.1322 &  \\
    \bottomrule
    \end{tabular}%
    }
  \caption{Sonar data: the two best settings obtained by averaging the \ecv values computed using BCV 4 x 4 and all CV partitioning strategies across different seeds.}
  \label{tab:so_2best}%
\end{table}%
\renewcommand{\baselinestretch}{\bsln}

As expected, the effect of the RF initial seeds is generally negligible. Table \ref{tab:so_randeffects} illustrates that, on this small data set, it accounts for substantially less variance compared to the CV partition. Nevertheless, it is not significantly different from zero only for the 5-Folds STS strategy. The columns labeled ``Prob'' show the empirical p-values calculated through permutation.
\renewcommand{\baselinestretch}{0.9}
\begin{table}[H]
  \centering
{\scriptsize
\begin{tabular}{lrrrrrrrc}
\toprule
      & \multicolumn{3}{c}{SRS} &       & \multicolumn{3}{c}{STS} &  \\
\cmidrule{2-4}\cmidrule{6-9}      & \multicolumn{1}{l}{Df} & \multicolumn{1}{l}{MSE} & \multicolumn{1}{l}{Prob} &       & \multicolumn{1}{l}{Df} & \multicolumn{1}{l}{MSE} & \multicolumn{1}{l}{Prob} &  \\
\cmidrule{1-4}\cmidrule{6-8}CVseeds & 3     & 0.0550 & 0     &       & 3     & 0.0809 & 0     & \multirow{4}[1]{*}{5-Folds} \\
repl:CVseeds & 3     & 0.0022 & 0     &       & 3     & 0.0027 & 0     &  \\
RFseeds & 3     & 0.0003 & 0.009 &       & 3     & 0.0001 & 0.726 &  \\
Total & 9     & 0.0192 &       &       & 9     & 0.0279 &       &  \\
      &       &       &       &       &       &       &       &  \\
CVseeds & 3     & 0.0927 & 0     &       & 3     & 0.0725 & 0     & \multirow{4}[1]{*}{10-Folds} \\
repl:CVseeds & 3     & 0.0026 & 0     &       & 3     & 0.0015 & 0     &  \\
RFseeds & 3     & 0.0005 & 0.001 &       & 3     & 0.0003 & 0.015 &  \\
Total & 9     & 0.0319 &       &       & 9     & 0.0248 &       &  \\
\bottomrule
\end{tabular}%
 }
  \caption{Sonar data: partial ANOVA tables comparing the MSE of only the random effects on the \ecv computed with BCV 4x4 for all four different partitioning strategies.}
  \label{tab:so_randeffects}%
\end{table}%
\renewcommand{\baselinestretch}{\bsln}
\subsection{Comparison with RCV}
Table \ref{tab:so_bcvvsrcv16} presents a comparison of the top five settings obtained using 5-BCV SRS 4x4 and 5-RCV SRS 16Rep. Similarly, Table \ref{tab:bcvvsrcv4} shows the same comparison for settings obtained using only four blocks or repetitions (5-BCV SRS 2x2 and 5-RCV SRS 4Rep). The results demonstrate that four and 16-block BCV yield comparable outcomes, while the results of RCV are less consistent with the others when only four repetitions are performed.
\renewcommand{\baselinestretch}{0.9}
\begin{table}[H]
  \centering
 {\scriptsize
\begin{tabular}{rrlrrrrrrlrrr}
\toprule
\multicolumn{6}{c}{BCV 4x4}                   &       & \multicolumn{6}{c}{RCV  16Rep} \\
\cmidrule{1-6}\cmidrule{8-13}\multicolumn{1}{l}{mtry} & \multicolumn{1}{l}{min.node} & repl  & \multicolumn{1}{l}{samp.frac} & \multicolumn{1}{l}{Sett.} & \multicolumn{1}{l}{$\ecvb$} &       & \multicolumn{1}{l}{mtry} & \multicolumn{1}{l}{min.node} & repl  & \multicolumn{1}{l}{samp.frac} & \multicolumn{1}{l}{Sett.} & \multicolumn{1}{l}{$\ecvb$} \\
\cmidrule{1-6}\cmidrule{8-13}5     & 3     & F     & 0.9   & 61    & 0.142 &       & 5     & 3     & F     & 0.9   & 61    & 0.147 \\
5     & 5     & F     & 0.9   & 64    & 0.144 &       & 5     & 5     & F     & 0.9   & 64    & 0.149 \\
5     & 3     & F     & 0.7   & 37    & 0.151 &       & 10    & 3     & F     & 0.9   & 62    & 0.157 \\
5     & 5     & F     & 0.7   & 40    & 0.154 &       & 10    & 5     & F     & 0.9   & 65    & 0.157 \\
10    & 3     & F     & 0.9   & 62    & 0.155 &       & 5     & 3     & F     & 0.7   & 37    & 0.159 \\
\bottomrule
\end{tabular}%
}
  \caption{Sonar data: comparison of the five settings with lowest $\ecvb$ computed with BCV 4x4 and  RCV 16 Rep.}
  \label{tab:so_bcvvsrcv16}%
\end{table}%
\renewcommand{\baselinestretch}{\bsln}

\begin{figure}[H]
  \centering
  \includegraphics[width=0.75\textwidth]{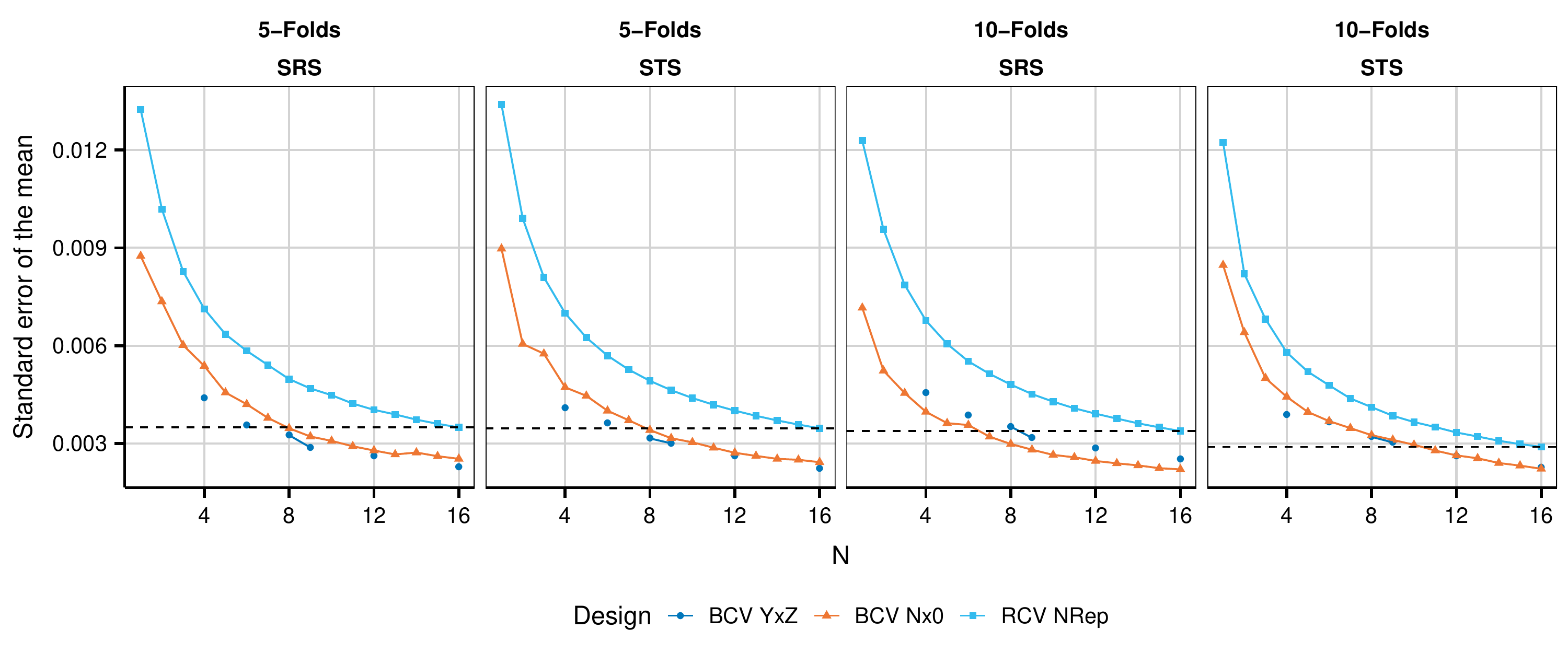}
  \caption{Sonar data: comparison of the standard error of $\ecvb$ for different number of runs and variance reduction designs. The horizontal lines mark the minimum std.err obtained with RCV.} \label{fig:so_stderr}%
\end{figure}

\renewcommand{\baselinestretch}{0.9}
\begin{table}[H]
  \centering
 {\scriptsize
\begin{tabular}{rrlrrrrrrlrrr}
\toprule
\multicolumn{6}{c}{BCV 2x2}                   &       & \multicolumn{6}{c}{RCV  4Rep} \\
\cmidrule{1-6}\cmidrule{8-13}\multicolumn{1}{l}{mtry} & \multicolumn{1}{l}{min.node} & repl  & \multicolumn{1}{l}{samp.frac} & \multicolumn{1}{l}{Sett.} & \multicolumn{1}{l}{$\ecvb$} &       & \multicolumn{1}{l}{mtry} & \multicolumn{1}{l}{min.node} & repl  & \multicolumn{1}{l}{samp.frac} & \multicolumn{1}{l}{Sett.} & \multicolumn{1}{l}{$\ecvb$} \\
\midrule
5     & 3     & F     & 0.9   & 61    & 0.154 &       & 10    & 5     & F     & 0.9   & 65    & 0.149 \\
5     & 5     & F     & 0.9   & 64    & 0.155 &       & 5     & 3     & F     & 0.9   & 61    & 0.150 \\
5     & 3     & T     & 1     & 73    & 0.157 &       & 10    & 3     & F     & 0.7   & 38    & 0.151 \\
5     & 5     & T     & 1     & 76    & 0.162 &       & 5     & 3     & F     & 0.7   & 37    & 0.155 \\
5     & 3     & F     & 0.7   & 37    & 0.163 &       & 5     & 3     & T     & 0.9   & 49    & 0.156 \\
\bottomrule
\end{tabular}%
}
  \caption{Sonar data: comparison of the five settings with lowest $\ecvb$computed with BCV 2x2 and  RCV 4 Rep.}
  \label{tab:bcvvsrcv4}%
\end{table}%
\renewcommand{\baselinestretch}{\bsln}
The standard errors (std.err) of $\ecvb$ are displayed for increasing numbers of blocks or repetitions (N) across different partitioning strategies in Figure \ref{fig:so_stderr}. The std.err for BCV YxZ is shown for specific configurations, such as 2x2, 3x2, 4x2, 3x3, 4x3, and 4x4, corresponding to 4, 6, 8, 9, 12, and 16 runs, respectively.

The results indicate that the std.errs obtained with BCV are consistently lower than the std.errs obtained with RCV for all numbers of runs. All curves exhibit a decreasing pattern at a rate of $\sqrt{N}$, in accordance with the variances of $\ecvb$ described later in equations \eqref{eq:bcvvar} and \eqref{eq:rcvvar}.
Notably, the std.errs of 10-BCV SRS YxZ are slightly higher compared to 10-BCV SRS Nx0. This increase in std.err can be attributed to the highly nonsignificant effect of the RF seeds, as illustrated, for example, in Table \ref{tab:so_bcvvsrcvanova}, which effectively transforms the design into a duplicated 2-block design.
Table \ref{tab:so_bcvvsrcvanova} presents a comparison of the ANOVA breakdown for three different designs: BCV 2x2 and 4x0, and RCV 16Rep, all computed with a 10-fold SRS strategy.
The results show that both BCV designs yield a significantly lower mean square error (MSE) for the effects and residuals compared to RCV 16Rep, even though they require one---fourth the number of runs. When blocking is done only with respect to the CV seeds (BCV 4x0), the residual MSE is lower than that of BCV 4x0, which leads to a lower std.err of the $\ecvb$.
 \renewcommand{\baselinestretch}{0.9}
\begin{table}[H]
  \centering
{\scriptsize
\begin{tabular}{llllrrrrrrrrrrr}
\toprule
      & \multicolumn{4}{c}{BCV 2x2}   &       & \multicolumn{4}{c}{BCV 4x0}   &       & \multicolumn{4}{c}{RCV 16Rep} \\
\cmidrule{1-5}\cmidrule{7-10}\cmidrule{12-15}source & \multicolumn{1}{r}{Df} & \multicolumn{1}{r}{$\text{SSE}^*$} & \multicolumn{1}{r}{$\text{MSE}^*$} & Prob  &       & Df    & $\text{SSE}^*$ & $\text{MSE}^*$ & Prob  &       &       &       &       &  \\
\cmidrule{1-5}\cmidrule{7-10}\multicolumn{10}{c}{\textbf{Random Effects}}                                  &       &       &       &       &  \\
CVseeds & \multicolumn{1}{r}{1} & \multicolumn{1}{r}{50.70} & \multicolumn{1}{r}{50.70} & 0     &       & 3     & 11.15 & 3.72  & 0.00  &       &       &       &       &  \\
repl:CVseeds & \multicolumn{1}{r}{1} & \multicolumn{1}{r}{3.26} & \multicolumn{1}{r}{3.26} & 0     &       & 3     & 0.90  & 0.30  & 0.00  &       &       &       &       &  \\
RFseeds & \multicolumn{1}{r}{1} & \multicolumn{1}{r}{0.02} & \multicolumn{1}{r}{0.02} & 1     &       &       &       &       &       &       &       &       &       &  \\
Total & \multicolumn{1}{r}{3} & \multicolumn{1}{r}{53.98} & \multicolumn{1}{r}{17.99} &       &       & 6     & 12.06 & 2.01  &       &       &       &       &       &  \\
\cmidrule{1-10}\multicolumn{15}{c}{\textbf{Fixed Effects}} \\
mtry  & \multicolumn{1}{r}{2} & \multicolumn{1}{r}{15.4} & \multicolumn{1}{r}{7.69} & 0     &       & 2     & 12.4  & 6.21  & 0     &       & 2     & 25.3  & 12.66 & 0 \\
min.node & \multicolumn{1}{r}{3} & \multicolumn{1}{r}{43.9} & \multicolumn{1}{r}{14.63} & 0     &       & 3     & 27.7  & 9.23  & 0     &       & 3     & 62.5  & 20.84 & 0 \\
replace & \multicolumn{1}{r}{1} & \multicolumn{1}{r}{2.3} & \multicolumn{1}{r}{2.27} & 0     &       & 1     & 2.5   & 2.50  & 0     &       & 1     & 15.7  & 15.66 & 0 \\
samp.frac & \multicolumn{1}{r}{3} & \multicolumn{1}{r}{24.8} & \multicolumn{1}{r}{8.27} & 0     &       & 3     & 27.2  & 9.06  & 0     &       & 3     & 92.1  & 30.70 & 0 \\
repl:samp.frac & \multicolumn{1}{r}{2} & \multicolumn{1}{r}{0.1} & \multicolumn{1}{r}{0.04} & 0.773 &       & 2     & 0.8   & 0.38  & 0.001 &       & 2     & 2.3   & 1.17  & 0 \\
Residuals & \multicolumn{1}{r}{322} & \multicolumn{1}{r}{30.0} & \multicolumn{1}{r}{0.09} &       &       & 318   & 20.1  & 0.06  &       &       & 1332  & 243.7 & 0.18  &  \\
\bottomrule
\multicolumn{4}{l}{$^*$ sum of squares values multiplied by 1000} &       &       &       &       &       &       &       &       &       &       &  \\
\end{tabular}%
  }
  \caption{Sonar data: ANOVA tables comparing 10--BCV SRS with two different types of blocking with 10--RCV SRS 16Rep.}
  \label{tab:so_bcvvsrcvanova}%
\end{table}%
\renewcommand{\baselinestretch}{\bsln}
\section{Related work}\label{sec:relwork}
There is a large body of literature on \ecv and learner tuning. Here, we synthesize a few results of interest. While all authors agree that \ecv has a bias and a variance, we are not aware of any theoretical works that explicitly discuss controlling for CV partitioning or algorithm's random behavior to reduce the \ecv variance when tuning a learner. Also, we did not find any discussion about the statistical model underpinning the use of \ecv for learner tuning with grid search. Furthermore, most articles are more concerned about the behaviour of \ecv as an estimate of the  generalized error, rather than on the training sample.


In their comprehensive overview, \cite{arl} discuss various CV techniques employed for tuning learners. They emphasize the significance of CV in achieving both accuracy and generalizability in learning algorithms. However, they caution that "no optimal CV method can be pointed out before having taken into account the final user's preferences." The authors acknowledge that the ideal number of folds typically falls within the range of five to ten. Nevertheless, they note that the variance of \ecv is influenced by several variables associated with the fold count, and in some cases, better statistical performance can be attained with a different number of folds. Additionally, the authors advise caution when utilizing stratified sampling, as it can significantly disrupt CV heuristics. We can confirm that, as we will show, stratified sampling does not always decrease the variance of \ecv.

In the context of choosing the number of folds for CV, several papers suggest employing 10-fold CV based on empirical evidence. For instance, \cite{mol} compared leave-one-out and 10-fold CV and found similar results, indicating that the former is more computationally efficient while maintaining accurate error estimation. Similarly, \cite{koh} compared prediction error estimates obtained from CV and bootstrap on small-sized data sets using classifiers C.4 and naive-bayes, and concluded that 10-fold stratified CV achieves a favorable bias-variance balance for model selection. Furthermore, \cite{kim} demonstrated that 10-fold CV exhibits lower variance compared to hold-out samples and bootstrap, while maintaining reasonable computational complexity. Lastly, \cite{kuh} recommend repeated 10-fold CV for small data sets due to its favorable bias and variance properties and manageable computational costs.



\cite{var} and \cite{bou} provide empirical evidence of a potential bias in estimating learner performance during parameter tuning, which arises from using different settings for \ecv computation. To mitigate this issue, \cite{bur} discusses the use of data transformation and a repeated mix of CV and hold-out samples to reduce both CV variance and bias, particularly in a regression context.

RCV (Repeated Cross-Validation) is a widely used technique in applications and textbooks, and it is implemented in various computer packages, including the R package "caret" \citep{kuhcar}. While the origin of RCV is uncertain, studies by \cite{mol} and \cite{kim} suggest that it can effectively enhance the precision of estimates while maintaining a small bias. Additionally, \cite{you21} found that \ecv computed with RCV exhibits smoothness compared to that computed with simple CV. However, \cite{van} acknowledge that RCV reduces the variance of \ecv but caution against its use due to the presence of bias.

Numerous authors have examined the variance of \ecv for different purposes. \cite{die} empirically demonstrated that statistical tests assuming a normal distribution can lead to a high probability of Type-I errors when comparing learners on the same data set using CV or a validation set, particularly for small data sets. \cite{bre84} proposed the "one-standard error" method for selecting simpler learners whose performance falls within one standard error of the best \ecv. Furthermore, \cite{efr83} utilized an ANOVA decomposition of \ecv based on orthogonal polynomials.

\cite{ber} proposed to replace grid search by randomly sampling hyperparameter values within predefined ranges and evaluating the learner's performance using each sampled configuration. By repeatedly sampling and evaluating different hyperparameter combinations, random search explores a wide range of possibilities in the hyperparameter space. It is difficult to cast this methodology into an ANOVA model, because the hyperparameters take different values in each setting. Since the \ecv s computed are unreplicated, they suffer from the high variance mentioned above and the results of the experiments may not be reliable. Nevertheless, random search is computationally more efficient than grid search and proves to be a practical solution, especially when dealing with a large number of hyperparameters.

In the context of identifying optimal settings, \cite{mor} suggested using advanced factorial designs and applying response surface methodology. This approach does not involve blocking for CV partitioning and is more suitable for sensitivity analysis purposes.
\section{A statistical model for grid search}\label{sec:theory}
Given a data sample $T = (X, Y)$, we consider a generic learning algorithm $f(X, \bth; R)$, with hyperparameters $\bth = (\theta_1, \theta_2,\ldots)$ and a random behavior $R$.
The CV error, denoted as $\ecvm(\bth_m; R, P)$, is obtained by applying the chosen loss function to the predictions made by the learning algorithm with hyperparameters $\bth_m = (\theta_{1m}, \theta_{2m},\ldots)$ and random behavior $R$, to each fold of a particular CV partition $P$.
We aim to estimate the expected \ecv for setting $\bth_m$ over the available data, denoted by $\ecvm(\bth_m)$, for a fixed strategy.
Since $\ecvm(\bth_m; R, P)$ is constant when $P$ and $R$ are fixed, the expectation over all possible values of \ecv in the sample $T$ is equivalent to:
\begin{equation}\label{eq:anovaCVnotheta}
\ecvm(\bth_m) = E\big[\ecvm(\bth_m; R, P)\big] = E_{R, P}\big[\ecvm(\bth_m; R, P)\vert R, P\big],
\end{equation}
where the subscripts under the expectation operator denote that the expectation is taken over all possible values of that variable.\footnote{In principle, we could compute $\ecvm(\bth_m)$ exactly by averaging over all possible  $\prod_{k = 1}^{K-1}\binom{n - (k - 1) n_t}{n_t}$  SRS K-fold partitions, where $n$ is the sample size and $n_t$ is the size of the folds (assumed constant for simplicity) and initial seeds for the learner. The number of possible seeds depends on the random number generator used, which easily reaches into the billions. In our case, the R default "Mersenne-Twister" random number generator uses a 32-bit integer as its seed value, resulting in $2^{32}$ or 4,294,967,296 possible seed values. Hence, considering all these possibilities, this computation is infeasible.}

Furthermore, also the variance of the \ecv is evaluated across CV partitions and random behaviors. Specifically, we have:
\begin{eqnarray}\label{eq:anovaCV}
&Var[\ecvm(\bth_m; R, P)] = E_{R, P}\big[Var(\ecvm(\bth_m; R, P)\vert R, P)\big] +\\ \nonumber\
&Var_{R, P}\big[E(\ecvm(\bth_m; R, P)\vert R, P)\big]
= Var_{R, P}\big[\ecvm(\bth_m; R, P)\vert R, P\big].
\end{eqnarray}
Henceforth, expectations will be considered as taken over CV partitions and random behaviors, and conditional to the training data available.
\subsection{Analysis of variance}
Experimental designs were first proposed in the early 19th century \citep{fis}, and have since been developed and refined. The general theory we present here can be found in any book on the subject, such as the classics \cite{box} or \cite{mon}.

A cartesian grid built from $M$ settings, $\bth_m$, and blocked with respect to $N_P$ CV seeds and $N_R$ initial seeds is a complete balanced block design with $M\times N_P\times N_R$ runs with $\ecvm(\bth_m; R, P)$ as response.

Experiments are typically analyzed by postulating an ANOVA linear model. For the blocked grid described above, the simplest such model is a first order mixed-effects model with random block effects and fixed setting effects, which can be written as
\begin{equation}\label{eq:anovamodel}
  \ecvm(\bth; R, P) = \mu + \pi + \rho + \tau + \vare,
\end{equation}
where $\mu$ is the overall mean \ecv, $\pi$ and $\rho$ are the random effects of the CV partition and random behavior, respectively, $\tau$ is the fixed effect of the settings and $\vare$ is an error term.  The random terms are supposed to be mutually independent and to have regular generic distributions with
\[
E(\pi) = E(\rho) = E(\vare) = 0,\, Var(\pi) = \sigma^2_\pi,\, Var(\rho) = \sigma^2_\rho,\, Var(\vare) = \sigma^2_\vare.
\]
Under this model, the expected value of $\ecvm(\bth_m; R, P)$ is given by:
\[
\ecvm(\bth_m) = E\big[\ecvm(\bth_m; R, P)\big] = \mu + \tau_m.
\]
Because of the independence assumptions, the variance of $\ecvm(\bth_m; R, P)$ is simply equal to the sum of the variances of the random terms, that is:
\[
  Var\big[\ecvm(\bth_m; R, P)\big] = Var(\pi) + Var(\rho) + Var(\vare).
\]

In contrast to the standard assumptions made in ANOVA, where the error term $\varepsilon$ in model \eqref{eq:anovamodel} represents an additional source of response variability, in our model the random components $R$ and $P$ are responsible for all the variability. Therefore, it is necessary to redefine the error term $\varepsilon$. Model \eqref{eq:anovamodel} is derived as a first-order MacLaurin approximation of $\ecvm(\bth; R, P)$. Consequently, $\varepsilon$ can be viewed as the sum of individually nonsignificant higher-order terms. This definition suits the broad goal of learner tuning. If there is evidence indicating that higher--order terms significantly differ from zero, they can be explicitly incorporated into the model. It is important to note that while the independence assumptions are employed to simplify the analysis, they may not be entirely satisfied.

The analysis of the results of blocked experiments is based on the discrete form of model \eqref{eq:anovamodel}:
\begin{equation}\label{eq:anovamodeldisc}
  \ebcvs{ijm} = \mu + \pi_i + \rho_j + \tau_m + e_{ijm},\ i = 1,\ldots, N_P,\ j = 1,\ldots, N_{R},
  \ m = 1,\ldots, M,
\end{equation}
where the indices refer to the different levels of each factor chosen for the experiments. The indices $i$ and $j$ correspond to different initial seeds for the CV partitioning and learner algorithm.
The setting effects, $\tau_m$, are broken down into the sum of the effects of the single hyperparameters, which can include higher order terms, such as interactions, for example
\[
\tau_m = \sum_t \gamma_t + \sum_t\sum_{s > t}\gamma_{st}.
\]
Since model \eqref{eq:anovamodeldisc} is not identified, different restrictions can be imposed to reduce the number of parameters. We adopt the zero-sum constraints, which require that
\[
\sum_j \pi_j = 0;\ \sum_k \gamma_k = 0;\ \sum_t \tau_t = 0.
\]
In a complete block balanced design, these constraints result in orthogonal contrasts and the
estimates of the block effects are the differences between the block or setting means and the overall mean. That is,
\[
\hat{\mu} = \ebcvbs{...}, \hat{\pi}_i = (\ebcvbs{i..} - \ebcvbs{...}),
\hat{\rho}_j = (\ebcvbs{.j.} - \ebcvbs{...}),\
 \hat{\tau}_m = (\ebcvbs{..m} - \ebcvbs{...}),
\]
where the "." in the subscript represents the averaging over the index it replaces and the residuals $\hat{e}_{ijm}$ are obtained by difference.
If the effects of individual hyperparameters are included in the model, their estimated are computed in a similar manner. Thus, the net estimate of $\ecvm(\bth)$ is the average over the random effects, that is:
\[
\widehat{Err^{CV}}\big(\bth_m\big) = \ebcvbs{..m} =
\widehat{\mu} + \widehat{\tau}_m.
\]
It can be proven that
\begin{equation}\label{eq:bcvvar}
E\big[\ebcvbs{..m}\big] = \mu + \tau_m;\ Var\big[\ebcvbs{..m}\big] =
\frac{\sigma_\vare^2}{N_PN_{R}}.
\end{equation}
In RCV the \ercv's are obtained by repeating $N$ times the grid, letting the seeds vary freely instead of blocking them. Model \eqref{eq:anovamodel} is still valid but, since the random effects are completely confounded with the error term, the empirical model becomes:
\begin{equation*}\label{eq:anovamodelRCVdisc}
  \ercvs{km} =  \mu +\tau_m + f_{km},\ k = 1\ldots, N,
  \ m = 1,\ldots, M.
\end{equation*}
The error term is equal to $f_{km} = \pi_{km} + \rho_{km}  + e_{km}$. The estimated \ercv for each setting is equal to the average over the repetitions, $\ercvbs{.m}$, which can be shown to have mean and variance equal to
\begin{equation}\label{eq:rcvvar}
E\big[\ercvbs{.m}\big] = \mu + \tau_m;\
Var\big[\ercvbs{.m}\big] =
\frac{\sigma^2_{\pi} + \sigma^2_{\rho} +
\sigma_\vare^2}{N}.
\end{equation}
Comparing the variance of the estimates obtained with BCV, equation \eqref{eq:bcvvar}, and with RCV, equation \eqref{eq:rcvvar}, it appears clear why blocking attains more precise estimates than randomization.

When a genuinely random error term is present, a design that is blocked with respect to a variable with a negligible variance is essentially equivalent to a repeated design. However, in the case of the response being the \ecv, where the sources of variability are solely attributed to $R$ and $P$, blocking with respect to one of these variables can be disadvantageous if its effect is negligible. This is because the \ecv for each level of the other blocking variable are essentially identical, making the two-block design almost a redundant copy of the one-block design. Consequently, this would lead to an inflation of the variance terms without providing any advantage. As mentioned earlier, the random behavior often has minimal impact on \ecv, rendering blocking with respect to $R$ ineffective or even counterproductive.
%
%
%
%
%
%
%
%
\subsection{Permutation tests}\label{sec:permtests}
In classical ANOVA, the estimated effects are tested by comparing the ratio of their MSE and the residual MSE to an F--distribution. However, in our setup, we cannot rely on F--tests because all of the variance of \ecv is accounted for by the random effects, and we do not want to make assumptions about the distribution of the errors.

To address this issue, permutation tests can be used as a nonparametric alternative. These tests rely on the weak exchangeability of the observations under the null hypotheses that one or more effects are zero, and they ignore the variance of the error. Therefore, they can be performed on dependent observations, provided that the exchangeability assumption is valid. A useful reference for permutation tests is provided by \cite{good}.

Permutation tests for blocked experimental designs can be challenging because of the restrictions on exchangeability introduced by the blocking factors. Various data transformations have been proposed to enable exchangeability for these tests \citep[see][for a concise overview and references]{fro}. On our balanced complete block designs with orthogonal contrasts, all of these data transformations reduce to the seminal method of \cite{ken}. This method effectively boils down to permuting the block's residuals, which are defined as the residuals with the block effects removed:
\[
r_{(ij)m} = \ebcvs{ijm} - \hat{\pi}_i - \hat{\rho}_j = \ebcvbs{..m} + \hat{e}_{ijm}.
\]
\section{Real Data Examples}\label{sec:dataanal}
We conducted tests on BCV using a variety of data sets, including small to medium sizes, as well as a larger data set. These data sets called for either classification or regression tasks, and in one case, both were applied. Specifications of these data sets are provided in Table \ref{tab:data sets}; the Sonar data set was extensively analyzed in Section \ref{seq:motex}. In accordance with our theoretical analysis, BCV consistently outperformed RCV across our computations, with the degree of improvement varying across data sets. Therefore, to maintain brevity and avoid redundancy, in this section we will present a selection of representative results. To allow an objective and unbiased evaluation of BCV, we have also included instances where the blocking of RF seeds was not significant and the improvement of BCV over RCV was marginal.
\renewcommand{\baselinestretch}{0.9}
\begin{table}[H]
  \centering
\scriptsize{
    \begin{tabular}{lrrlrrrr}
    \toprule
          &       &       & \multicolumn{2}{c}{Task} &       & \multicolumn{2}{c}{ cpu time BCV 4x4} \\
\cmidrule{4-8}    Data set & \multicolumn{1}{l}{Cases} & \multicolumn{1}{l}{Features} & Class & \multicolumn{1}{l}{Regr} &       & 5-CV  & 10-CV \\
    \midrule
    Sonar & 208   & 60    & \checkmark &       &       & 2M 27S & 3M 40S \\
    Penguins & 342   & 5     & \checkmark &       &       & 2M 05S & 3M 31S \\
    Boston & 506   & 13    &       & \multicolumn{1}{l}{\checkmark} &       & 2M 55S & 4M 17S \\
    Insurance & 1338  & 6     &       & \multicolumn{1}{l}{\checkmark} &       & 3M 27S & 5M 48S \\
    Wine  & 4898  & 11    & \checkmark & \multicolumn{1}{l}{\checkmark} &       & 4M 33S & 7M 14S \\
    Adults & 30162 & 13    & \checkmark &       &       & 50M 50S & 1H 59M 25S \\
    \bottomrule
    \end{tabular}%
    }
    \caption{Data sets used in the examples}
\label{tab:data sets}%
\end{table}%
\renewcommand{\baselinestretch}{\bsln}

All the examples below were obtained using a cartesian grid of 84 settings, blocked with different number of randomly chosen initial seeds for CV partitioning and RF algorithm. The \ecv was computed by running RFs with 2,000 trees, unless otherwise specified. For other details, such as grid construction, and sampling and variance reduction designs notation, please refer to Section \ref{seq:motex}.

The computations were performed using the \textsf{R} language \citep{R} with our own implementation of BCV and RCV. Random Forests were executed using the \textsf{ranger} package \citep{ranger}, and permutation tests were conducted using the \textsf{lmPerm} package \citep{lmPerm}. All computation times were measured on a quad-core Xeon 3.6GHz CPU with 32GB of RAM.



\subsection{Penguin Data}
The Palmer penguins \citep{Penguin} data is a well known data set often used in didactic examples. It contains five characteristics of penguins of three species. The task is to classify the birds into the right species.
The classification task is very simple and usually learners achieve a very low classification error.

We computed the \ecv values on a grid defined by the hyperparameters values shown in Table \ref{tab:pe_parlist}.
\renewcommand{\baselinestretch}{0.9}
\begin{table}[H]
  \centering
{\scriptsize
    \begin{tabular}{ll}
    \toprule
    Hyperparameter & Values \\
    \midrule
    mtry  & 2, 3, 5 \\
    min.node.size & 1,  3,  5, 10 \\
    replace & T, F \\
    sample.fraction & 0.5, 0.7, 0.9, 1.0 \\
    \bottomrule
    \end{tabular}%
    }
  \caption{Penguin data: hyperparameter values used for tuning.}
  \label{tab:pe_parlist}%
\end{table}%
\renewcommand{\baselinestretch}{\bsln}
Table \ref{tab:pe_Best2} compares the settings corresponding to the two lowest $\ecvb$ for the four different partitioning strategies, obtained with BCV 4x4 and 16--RCV. There is a substantial agreement among the best settings identified by the different designs and sampling strategies. Because of the simplicity of the task, many different settings yield low \ecv s that are almost identical within strategies. All best settings have $mtry = 2$ and sampling with replacement, large $sample.fraction$ and $min.node.size$ equal to one or three.
\renewcommand{\baselinestretch}{0.9}
\begin{table}[H]
  \centering
{\scriptsize
\begin{tabular}{rrlrrrrrrlrrrrl}
\cmidrule{1-6}\cmidrule{8-13}\cmidrule{15-15}\multicolumn{6}{c}{BCV 4x4}                   &       & \multicolumn{6}{c}{RCV 16Rep}                 &       &  \\
\cmidrule{1-6}\cmidrule{8-13}\cmidrule{15-15}\multicolumn{1}{l}{mtry} & \multicolumn{1}{l}{min.nod} & repl. & \multicolumn{1}{l}{samp.frac} & \multicolumn{1}{l}{Sett.} & \multicolumn{1}{l}{$\ecvb$} &       & \multicolumn{1}{l}{mtry} & \multicolumn{1}{l}{min.nod} & repl. & \multicolumn{1}{l}{samp.frac} & \multicolumn{1}{l}{Sett.} & \multicolumn{1}{l}{$\ecvb$} &       & Strategy \\
\cmidrule{1-6}\cmidrule{8-13}\cmidrule{15-15}2     & 1     & T     & 0.9   & 49    & 0.0143 &       & 2     & 1     & T     & 1     & 73    & 0.0133 &       & \multicolumn{1}{c}{\multirow{2}[2]{*}{Overall}} \\
2     & 3     & T     & 0.9   & 52    & 0.0143 &       & 2     & 3     & T     & 1     & 76    & 0.0135 &       &  \\
\cmidrule{1-6}\cmidrule{8-13}\cmidrule{15-15}2     & 1     & T     & 1     & 73    & 0.0155 &       & 2     & 1     & T     & 1     & 73    & 0.0133 &       & \multirow{2}[2]{*}{5-F SRS} \\
2     & 3     & T     & 1     & 76    & 0.0159 &       & 2     & 1     & F     & 0.9   & 61    & 0.0139 &       &  \\
\cmidrule{1-6}\cmidrule{8-13}\cmidrule{15-15}2     & 3     & T     & 0.9   & 52    & 0.0144 &       & 2     & 1     & T     & 0.9   & 49    & 0.0130 &       & \multirow{2}[2]{*}{5-F STS} \\
2     & 1     & T     & 0.9   & 49    & 0.0148 &       & 2     & 3     & T     & 1     & 76    & 0.0130 &       &  \\
\cmidrule{1-6}\cmidrule{8-13}\cmidrule{15-15}2     & 1     & T     & 0.7   & 25    & 0.0132 &       & 2     & 1     & T     & 1     & 73    & 0.0122 &       & \multirow{2}[2]{*}{10-F SRS} \\
2     & 1     & T     & 1     & 73    & 0.0132 &       & 2     & 1     & T     & 0.9   & 49    & 0.0132 &       &  \\
\cmidrule{1-6}\cmidrule{8-13}\cmidrule{15-15}2     & 1     & T     & 0.9   & 49    & 0.0126 &       & 2     & 1     & T     & 0.9   & 49    & 0.0126 &       & \multirow{2}[2]{*}{10-F STS} \\
2     & 3     & T     & 0.9   & 52    & 0.0126 &       & 2     & 1     & T     & 1     & 73    & 0.0128 &       &  \\
\cmidrule{1-6}\cmidrule{8-13}\cmidrule{15-15}\end{tabular}%
  }
   \caption{Penguin data: the two settings that yielded the lowest $\ecvb$ for BCV 4x4 and RCV 16Rep, using all partitioning strategies.}
 \label{tab:pe_Best2}%
\end{table}%
\renewcommand{\baselinestretch}{\bsln}

\begin{figure}[H]
  \centering
  \includegraphics[width=0.95\textwidth]{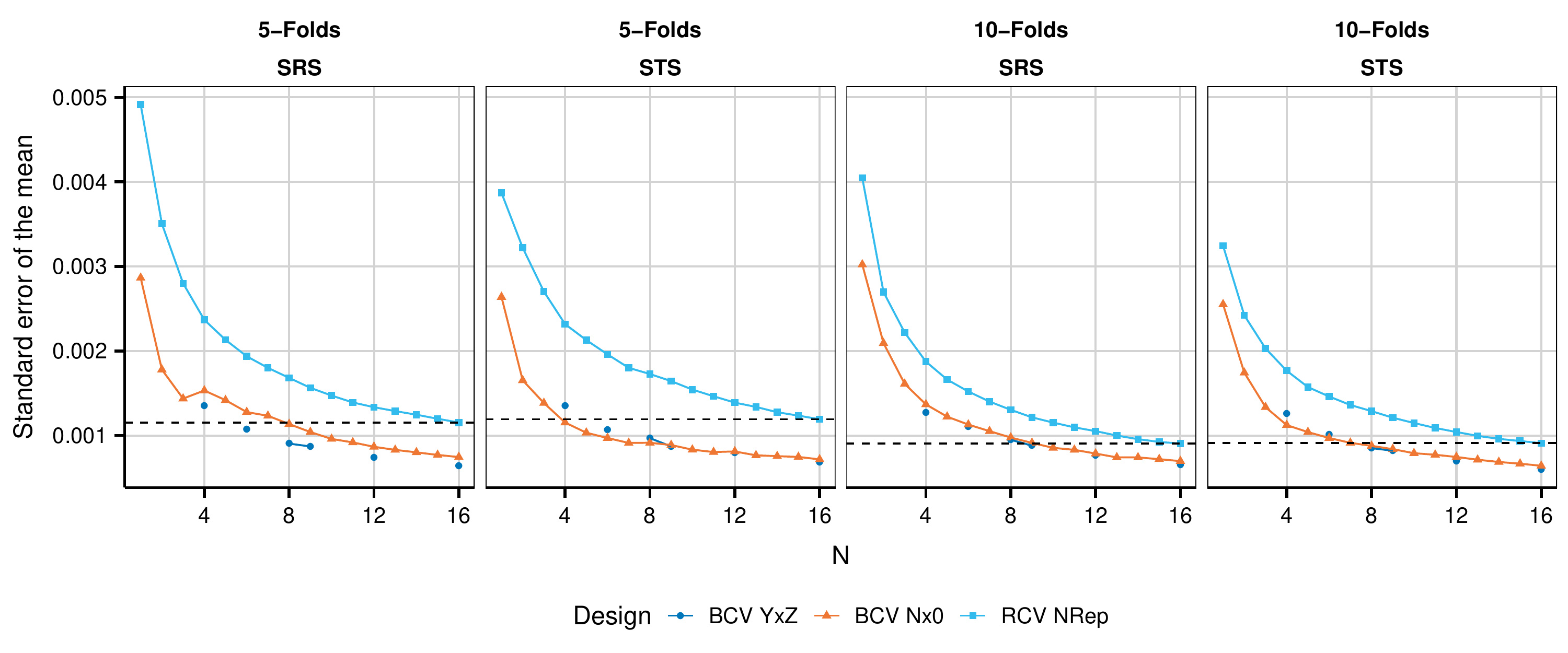}
  \caption{Penguin data: comparison of the standard error of $\ecvb$ for different number of runs and variance reduction designs. The horizontal lines mark the minimum std.err obtained with RCV.} \label{fig:pe_stderr}%
\end{figure}

In Figure \ref{fig:pe_stderr} it is evident that the std.err achieved with BCV consistently outperforms that obtained with RCV across increasing numbers of repetitions. BCV achieves a lower std.err with about half the number of runs.
The std.err of BCV YxZ is slightly higher than that of BCV Nx0 when employing 5-fold sampling and STS. This observation can be attributed to the highly nonsignificant effect of the RF seeds.
\renewcommand{\baselinestretch}{0.9}
\renewcommand{\baselinestretch}{\bsln}
Table \ref{tab:pe_aov_comp 4CVvs16rep} shows that the residual and hyperparameters MSEs calculated by blocking with only four CVseeds are much smaller than that obtained with 16 repetitions. IN the BCV table, the MSE of the block effects is comparatively large when compared to that of the hyperparameters, except for $mtry$.
\renewcommand{\baselinestretch}{0.9}
\begin{table}[H]
  \centering
{\scriptsize
\begin{tabular}{llllllrrrr}
\toprule
      & \multicolumn{4}{c}{BCV STS, 4x0} &       & \multicolumn{4}{c}{RCV 16 rep} \\
\cmidrule{2-5}\cmidrule{7-10}source & \multicolumn{1}{r}{Df} & \multicolumn{1}{r}{$\text{SSE}^*$} & \multicolumn{1}{r}{$\text{MSE}^*$} & \multicolumn{1}{r}{Prob} &       & Df    & $\text{SSE}^*$ & $\text{MSE}^*$ & Prob \\
\cmidrule{1-5}\cmidrule{7-10}\multicolumn{5}{c}{Random Effects}    &       &       &       &       &  \\
CVseeds & \multicolumn{1}{r}{3} & \multicolumn{1}{r}{2.51} & \multicolumn{1}{r}{0.8374} & \multicolumn{1}{r}{0} &       &       &       &       &  \\
repl:CVseeds         & \multicolumn{1}{r}{3} & \multicolumn{1}{r}{0.02} & \multicolumn{1}{r}{0.0065} & \multicolumn{1}{r}{0.78} &       &       &       &       &  \\
Total & \multicolumn{1}{r}{6} & \multicolumn{1}{r}{2.53} & \multicolumn{1}{r}{0.4219} &       &       &       &       &       &  \\
\midrule
\multicolumn{10}{c}{Fixed Effects} \\
mtry                    & \multicolumn{1}{r}{2} & \multicolumn{1}{r}{4.83} & \multicolumn{1}{r}{2.413} & \multicolumn{1}{r}{0} &       & 2     & 36.77 & 18.384 & 0 \\
min.node.size           & \multicolumn{1}{r}{3} & \multicolumn{1}{r}{2.66} & \multicolumn{1}{r}{0.888} & \multicolumn{1}{r}{0} &       & 3     & 9.35  & 3.118 & 0 \\
replace                 & \multicolumn{1}{r}{1} & \multicolumn{1}{r}{0.00} & \multicolumn{1}{r}{0.003} & \multicolumn{1}{r}{0.66} &       & 1     & 0.26  & 0.260 & 0 \\
sample.frac        & \multicolumn{1}{r}{3} & \multicolumn{1}{r}{0.36} & \multicolumn{1}{r}{0.120} & \multicolumn{1}{r}{0} &       & 3     & 0.50  & 0.167 & 0 \\
repl:sample.frac & \multicolumn{1}{r}{2} & \multicolumn{1}{r}{0.00} & \multicolumn{1}{r}{0.000} & \multicolumn{1}{r}{1} &       & 2     & 0.16  & 0.079 & 0 \\
Residuals               & \multicolumn{1}{r}{318} & \multicolumn{1}{r}{2.52} & \multicolumn{1}{r}{0.008} &       &       & 1332  & 17.64 & 0.013 &  \\
\midrule
\multicolumn{6}{l}{$^*$  values multiplied by 1,000} &       &       &       &  \\
\end{tabular}%
    }
  \caption{Penguin data: ANOVA tables for BCV 4x0 and RCV 16Rep obtained with 10--folds and STS strategy.}
  \label{tab:pe_aov_comp 4CVvs16rep}%
\end{table}%
\renewcommand{\baselinestretch}{\bsln}

\subsection{Boston Crime data}
The task for this data set \citep{Boston} is to predict the percentage rate of crime in 506 suburbs using 13 housing features. We computed the \ecv values on a grid defined by the hyperparameters values shown in Table \ref{tab:bo_parlist}.
\renewcommand{\baselinestretch}{0.9}
\begin{table}[H]
  \centering
{\scriptsize
\begin{tabular}{ll}
\toprule
Hyperparameter & Values \\
\midrule
mtry  & 3  7 11 \\
min.node.size & 3  5 10 15 \\
replace & T F \\
sample.fraction &  0.5 0.7 0.9 1.0 \\
\bottomrule
\end{tabular}%
    }
  \caption{Boston data: hyperparameter values used for tuning.}
  \label{tab:bo_parlist}%
\end{table}%
\renewcommand{\baselinestretch}{\bsln}
Table \ref{tab:bo_Best2} shows the two best settings found using BCV with eight blocks (BCV 4x2 and BCV 8x0), and RCV 16Rep. The values of $\ecvb$ are comparable and there is substantial agreement among the optimal settings, even though RCV tends to indicate different optimal settings for \textit{replace} and \textit{sample.fraction}.

\begin{figure}[H]
  \centering
  \includegraphics[width=0.75\textwidth]{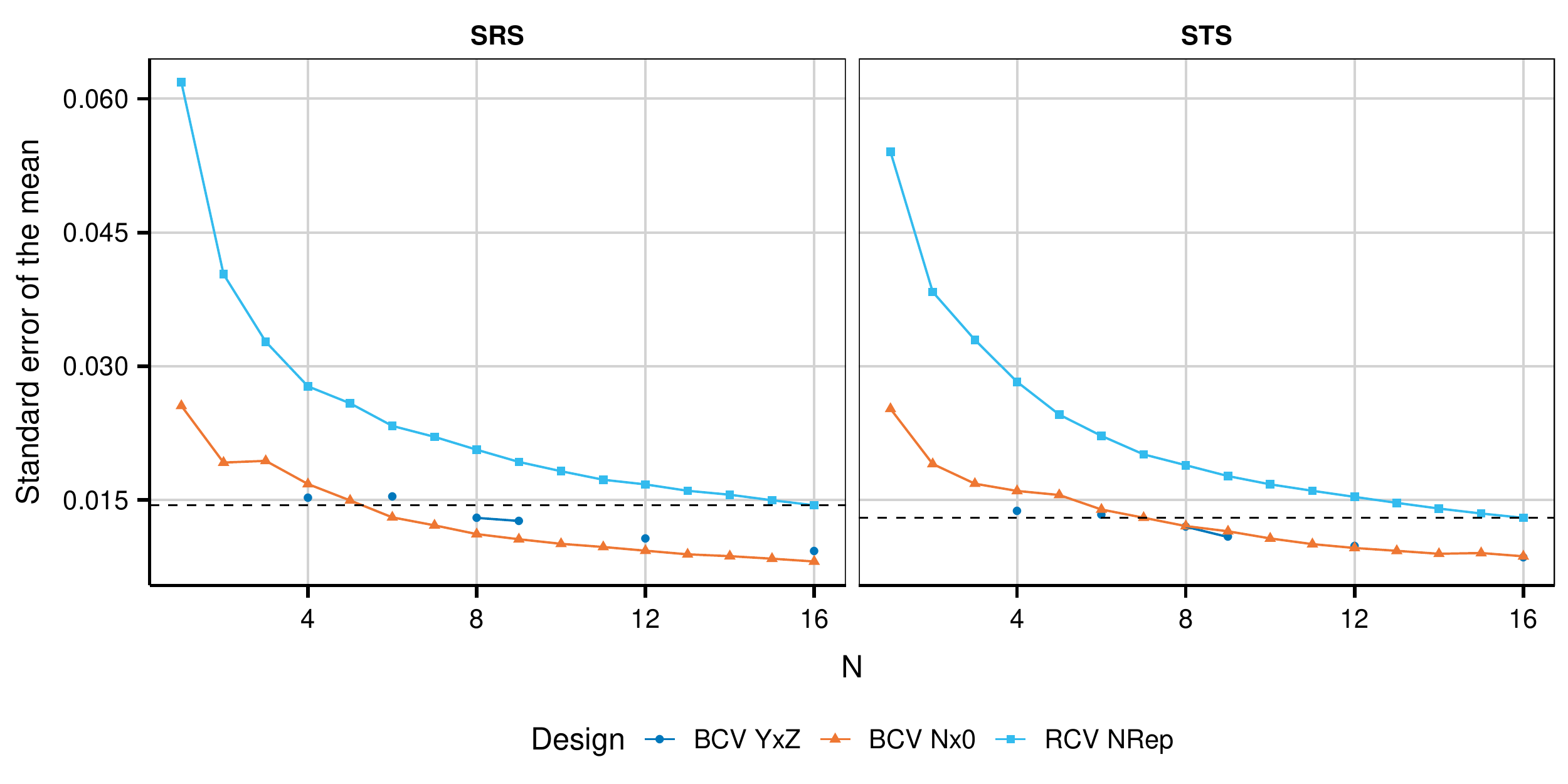}
  \caption{Boston data: comparison of std.err computed using 10-Fold and SRS or STS strategies and varying numbers of runs. The horizontal lines mark the minimum std.err obtained with RCV.} \label{fig:bo_stderr}%
\end{figure}

Figure \ref{fig:bo_stderr} shows that BCV consistently outperforms RCV in terms of std.err across all repetitions. Interestingly, BCV achieves lower std.err with approximately half the number of runs. Additionally, when using 10-fold SRS sampling for this data set, BCV YxZ yields slightly higher std.err than BCV Yx0, likely due to the nonsignificant effect of the RFseeds, as shown in Table \ref{tab:bo_ANOVA3}.
\renewcommand{\baselinestretch}{0.9}
\begin{table}[H]
  \centering
{\scriptsize
\begin{tabular}{rrlrrrl}
\toprule
\multicolumn{1}{l}{mtry} & \multicolumn{1}{l}{min.node} & repl. & \multicolumn{1}{l}{samp.frac.} & \multicolumn{1}{l}{Sett.} & \multicolumn{1}{l}{$\ecvb$} & Design \\
\midrule
3     & 15    & F     & 0.9   & 70    & 1.757 & \multirow{2}[2]{*}{Overall 10--BCV} \\
3     & 10    & F     & 0.7   & 43    & 1.760 &  \\
\midrule
\multicolumn{7}{c}{SRS} \\
\midrule
3     & 15    & F     & 0.9   & 70    & 1.745 & \multirow{2}[2]{*}{BCV 4x2} \\
3     & 15    & T     & 0.9   & 58    & 1.746 &  \\
\midrule
3     & 15    & F     & 0.9   & 70    & 1.746 & \multirow{2}[2]{*}{BCV 8x0} \\
3     & 10    & F     & 0.7   & 43    & 1.746 &  \\
\midrule
3     & 15    & T     & 0.9   & 58    & 1.752 & \multirow{2}[2]{*}{RCV 16Rep} \\
3     & 15    & T     & 1     & 82    & 1.753 &  \\
\midrule
\multicolumn{7}{c}{STS} \\
\midrule
3     & 15    & F     & 0.7   & 46    & 1.771 & \multirow{2}[2]{*}{BCV 4x2} \\
3     & 10    & F     & 0.5   & 19    & 1.772 &  \\
\midrule
3     & 15    & F     & 0.7   & 46    & 1.755 & \multirow{2}[2]{*}{BCV 8x0} \\
3     & 10    & F     & 0.7   & 43    & 1.756 &  \\
\midrule
3     & 15    & F     & 0.9   & 70    & 1.738 & \multirow{2}[2]{*}{RCV 16Rep} \\
3     & 10    & T     & 0.9   & 55    & 1.744 &  \\
\bottomrule
\end{tabular}%
  }
  \caption{Boston data: the two settings that yielded the lowest \protect$\ecvb$ for BCV 4x2, BCV 8x0 and RCV 16Rep, sampling with and without stratification, and 10-folds.}%
  \label{tab:bo_Best2}%
\end{table}%
\renewcommand{\baselinestretch}{\bsln}

The ANOVA tables presented in Table \ref{tab:bo_ANOVA3} shows how both BCV 4x2 and BCV 4x0 yield a lower MSE than RCV 16Rep, using 10--folds and SRS strategy.
\renewcommand{\baselinestretch}{0.9}
\begin{table}[H]
  \centering
{\scriptsize
\begin{tabular}{lrrrrrrrrrrrrrr}
\toprule
      & \multicolumn{4}{c}{BCV 4x2} &       & \multicolumn{4}{c}{BCV 4x0}   &       & \multicolumn{4}{c}{RCV 16Rep} \\
\cmidrule{2-5}\cmidrule{7-10}\cmidrule{12-15}source & \multicolumn{1}{l}{Df} & \multicolumn{1}{l}{SSE$^*$} & \multicolumn{1}{l}{MSE$^*$} & \multicolumn{1}{l}{Prob} &       & \multicolumn{1}{l}{Df} & \multicolumn{1}{l}{SSE$^*$} & \multicolumn{1}{l}{MSE$^*$} & \multicolumn{1}{l}{Prob} &       & \multicolumn{1}{l}{Df} & \multicolumn{1}{l}{SSE$^*$} & \multicolumn{1}{l}{MSE$^*$} & \multicolumn{1}{l}{Prob} \\
\cmidrule{1-5}\cmidrule{7-10}\cmidrule{12-15}\multicolumn{10}{c}{{Random Effects}}                                  &       &       &       &       &  \\
\cmidrule{1-10}CVseeds & 3     & 21.2  & 7.06  & 0     &       & 3     & 2.5   & 0.83  & 0     &       &       &       &       &  \\
repl:CV & 3     & 4.6   & 1.52  & 0     &       & 3     & 0.6   & 0.21  & 0.17  &       &       &       &       &  \\
RFseeds & 1     & 0.1   & 0.05  & 0.22  &       &       &       &       &       &       &       &       &       &  \\
Total & 7     & 25.8  &       &       &       & 6     & 3.1   &       &       &       &       &       &       &  \\
\midrule
\multicolumn{15}{c}{{Fixed  Effects}} \\
mtry                    & 2     & 84.0  & 42.00 & 0     &       & 2     & 58.2  & 29.08 & 0     &       & 2     & 233.2 & 116.60 & 0 \\
min.node           & 3     & 10.0  & 3.32  & 0     &       & 3     & 2.4   & 0.80  & 0     &       & 3     & 28.8  & 9.61  & 0 \\
replace                 & 1     & 55.2  & 55.19 & 0     &       & 1     & 26.1  & 26.05 & 0     &       & 1     & 120.0 & 120.00 & 0 \\
sample.frac      & 3     & 65.4  & 21.81 & 0     &       & 3     & 28.2  & 9.40  & 0     &       & 3     & 151.6 & 50.55 & 0 \\
repl:smp.fr & 2     & 26.6  & 13.30 & 0     &       & 2     & 10.4  & 5.20  & 0     &       & 2     & 50.1  & 25.05 & 0 \\
Residuals               & 653   & 92.7  & 0.14  &       &       & 318   & 32.5  & 0.10  &       &       & 1332  & 442.9 & 0.33  &  \\
\bottomrule
\multicolumn{5}{l}{$^*$ sum of squares values multiplied by 1000} &       &       &       &       &       &       &       &       &       &  \\
\end{tabular}%
 }
   \caption{Boston data:  ANOVA tables comparing BCV 4x2, BCV 4x0 and RCV 16Rep, using 10--CV SRS strategy.}
 \label{tab:bo_ANOVA3}%
\end{table}%
\renewcommand{\baselinestretch}{\bsln}

\subsection{Insurance data}
The Insurance data set \citep{Insurance} require predicting 1338 instances of medical costs billed by health insurance companies using six beneficiary's characteristics. We computed the \ecv values on a grid defined by the hyperparameters values shown in Table \ref{tab:ic_param}.

\renewcommand{\baselinestretch}{0.9}
\begin{table}[H]
  \centering
{\scriptsize
\begin{tabular}{ll}
\toprule
Hyperparameter & Values \\
\midrule
mtry  & 2 4 6 \\
min.node.size & 3  5 10 15 \\
replace & T F \\
sample.fraction & 0.5 0.7 0.9 1.0 \\
\bottomrule
\end{tabular}%
 }
  \caption{Insurance data: hyperparameter values used for tuning.}
  \label{tab:ic_param}%
\end{table}%
\renewcommand{\baselinestretch}{\bsln}
Table \ref{tab:ic_best2} shows that both BCV with four or eight blocks yield the same optimal settings, which closely resemble those obtained by RCV 16Rep. The $\ecvb$ values are also comparable across these approaches.
\renewcommand{\baselinestretch}{0.9}
\begin{table}[H]
  \centering
{\scriptsize
\begin{tabular}{llllrrr}
\toprule
mtry  & min.node & repl. & samp.frac. & \multicolumn{1}{l}{Sett.} & \multicolumn{1}{l}{$\ecvb^*$} & \multicolumn{1}{l}{Design} \\
\midrule
\multicolumn{1}{r}{4} & \multicolumn{1}{r}{15} & T     & \multicolumn{1}{r}{0.5} & 11    & 2070.7 & \multicolumn{1}{l}{\multirow{2}[2]{*}{Overall 10--BCV}} \\
\multicolumn{1}{r}{4} & \multicolumn{1}{r}{15} & F     & \multicolumn{1}{r}{0.5} & 23    & 2079.2 &  \\
\midrule
\multicolumn{7}{c}{SRS} \\
\midrule
\multicolumn{1}{r}{4} & \multicolumn{1}{r}{15} & T     & \multicolumn{1}{r}{0.5} & 11    & 2070.3 & \multicolumn{1}{l}{\multirow{2}[2]{*}{BCV 4x2}} \\
\multicolumn{1}{r}{4} & \multicolumn{1}{r}{15} & F     & \multicolumn{1}{r}{0.5} & 23    & 2078.5 & \multicolumn{1}{l}{} \\
\midrule
\multicolumn{1}{r}{4} & \multicolumn{1}{r}{15} & T     & \multicolumn{1}{r}{0.5} & 11    & 2072.8 & \multicolumn{1}{l}{\multirow{2}[2]{*}{BCV 4x0}} \\
\multicolumn{1}{r}{4} & \multicolumn{1}{r}{15} & F     & \multicolumn{1}{r}{0.5} & 23    & 2080.2 & \multicolumn{1}{l}{} \\
\midrule
\multicolumn{1}{r}{4} & \multicolumn{1}{r}{15} & T     & \multicolumn{1}{r}{0.5} & 11    & 2072.7 & \multicolumn{1}{l}{\multirow{2}[2]{*}{RCV 16Rep}} \\
\multicolumn{1}{r}{4} & \multicolumn{1}{r}{10} & T     & \multicolumn{1}{r}{0.5} & 8     & 2080.4 &  \\
\midrule
\multicolumn{7}{c}{STS} \\
\midrule
\multicolumn{1}{r}{4} & \multicolumn{1}{r}{15} & T     & \multicolumn{1}{r}{0.5} & 11    & 2070.6 & \multicolumn{1}{l}{\multirow{2}[2]{*}{BCV 4x2}} \\
\multicolumn{1}{r}{4} & \multicolumn{1}{r}{15} & F     & \multicolumn{1}{r}{0.5} & 23    & 2078.4 & \multicolumn{1}{l}{} \\
\midrule
\multicolumn{1}{r}{4} & \multicolumn{1}{r}{15} & T     & \multicolumn{1}{r}{0.5} & 11    & 2073.2 & \multicolumn{1}{l}{\multirow{2}[2]{*}{BCV 4x0}} \\
\multicolumn{1}{r}{4} & \multicolumn{1}{r}{15} & F     & \multicolumn{1}{r}{0.5} & 23    & 2079.1 & \multicolumn{1}{l}{} \\
\midrule
\multicolumn{1}{r}{4} & \multicolumn{1}{r}{15} & T     & \multicolumn{1}{r}{0.5} & 11    & 2074.3 & \multicolumn{1}{l}{\multirow{2}[2]{*}{RCV 16Rep}} \\
\multicolumn{1}{r}{4} & \multicolumn{1}{r}{15} & F     & \multicolumn{1}{r}{0.5} & 23    & 2081.2 &  \\
\midrule
\multicolumn{4}{l}{$^*$ values divided by 1,000} &       &       &  \\
\end{tabular}%
 }
   \caption{Insurance data: the two settings that yielded the lowest $\ecvb$ for BCV 4x4, BCV 4x0 and RCV 16Rep, using 10--fold with SRS and STS partitioning strategies.}
 \label{tab:ic_best2}%
\end{table}%
\renewcommand{\baselinestretch}{\bsln}
Table \ref{tab:ic_anova} presents the ANOVA tables for BCV 4x2, BCV 4x0, and RCV Rep16 using 10-folds and SRS strategy. These tables reveal that the MSEs of the effects achieved with four and eight blocks are lower than those obtained with a larger number of repetitions. Importantly, the effect of the RF seed is highly nonsignificant, resulting in similar outcomes between BCV 4x0 and BCV 4x2.

\begin{figure}[H]
  \centering
  \includegraphics[width=0.75\textwidth]{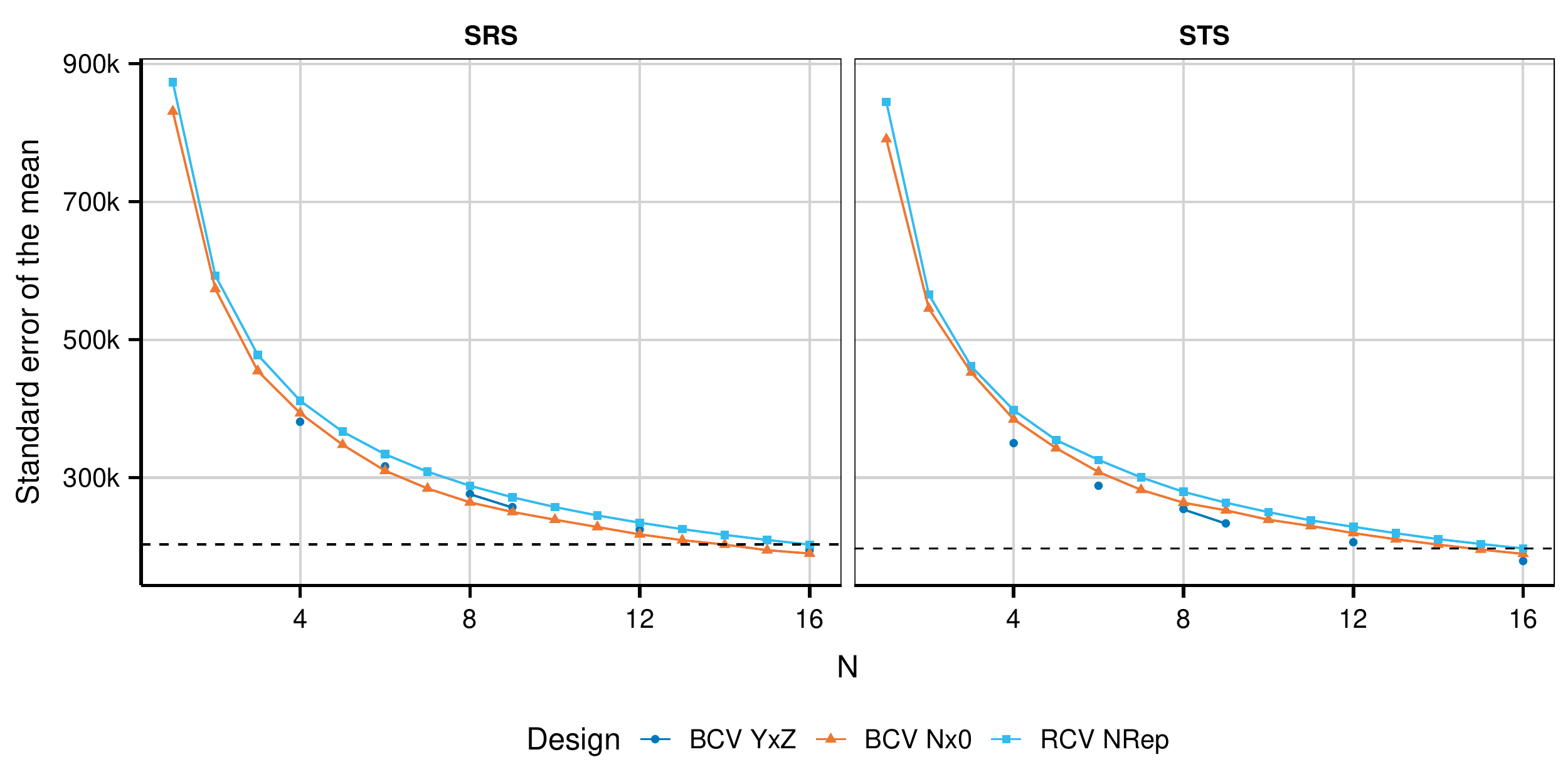}
  \caption{Insurance data: comparison of std.err for different run numbers and variance reduction designs using 10-fold STS and SRS. Values in 1,000\$. Horizontal lines indicate the minimum std.err obtained with RCV.} \label{fig:ic_stderr}%
\end{figure}

Figure \ref{fig:ic_stderr} depicts the consistent superiority of BCV over RCV in terms of std.err for this data set. Notably, the plotted std.err values are in 1000\$, so even seemingly small differences in the plots correspond to substantial differences in monetary terms. For instance, the std.errs for 10-Folds STS BCV 4x4, BCV 16x0, and RCV 16Rep are 178,408.4\$, 189,470.7\$, and 197,267.7\$, respectively.
\renewcommand{\baselinestretch}{0.9}
\begin{table}[H]
  \centering
{\scriptsize
\begin{tabular}{llllrrrrrrrrrrr}
\toprule
      & \multicolumn{4}{c}{BCV 4x2}   &       & \multicolumn{4}{c}{BCV 4x0}   &       & \multicolumn{4}{c}{RCV 16 Rep} \\
\cmidrule{2-5}\cmidrule{7-10}\cmidrule{12-15}source & Df    & SSE$^*$ & MSE$^*$ & \multicolumn{1}{l}{Prob} &       & \multicolumn{1}{l}{Df} & \multicolumn{1}{l}{SSE$^*$} & \multicolumn{1}{l}{MSE$^*$} & \multicolumn{1}{l}{Prob} &       & \multicolumn{1}{l}{Df} & \multicolumn{1}{l}{SSE$^*$} & \multicolumn{1}{l}{MSE$^*$} & \multicolumn{1}{l}{Prob} \\
\cmidrule{1-5}\cmidrule{7-10}\cmidrule{12-15}\multicolumn{10}{c}{Random Effects}                                           &       &       &       &       &  \\
CVseeds                 & \multicolumn{1}{r}{3} & \multicolumn{1}{r}{18.3} & \multicolumn{1}{r}{6.09} & 0     &       & 3     & 5.7   & 1.91  & 0.01  &       &       &       &       &  \\
repl:CVseeds         & \multicolumn{1}{r}{3} & \multicolumn{1}{r}{0.5} & \multicolumn{1}{r}{0.15} & 1     &       & 3     & 0.3   & 0.11  & 0.85  &       &       &       &       &  \\
RFseeds                 & \multicolumn{1}{r}{1} & \multicolumn{1}{r}{0.0} & \multicolumn{1}{r}{0.03} & 1     &       &       &       &       &       &       &       &       &       &  \\
Total & \multicolumn{1}{r}{7} & \multicolumn{1}{r}{18.7} & \multicolumn{1}{r}{2.68} &       &       & 6     & 6.1   & 1.01  &       &       &       &       &       &  \\
\midrule
\multicolumn{15}{c}{Fixed Effects} \\
mtry                    & \multicolumn{1}{r}{2} & \multicolumn{1}{r}{146.8} & \multicolumn{1}{r}{73.39} & 0     &       & 2     & 80.5  & 40.24 & 0     &       & 2     & 285.8 & 142.90 & 0 \\
min.node & \multicolumn{1}{r}{3} & \multicolumn{1}{r}{34.2} & \multicolumn{1}{r}{11.40} & 0     &       & 3     & 19.7  & 6.57  & 0     &       & 3     & 73.8  & 24.61 & 0 \\
replace                 & \multicolumn{1}{r}{1} & \multicolumn{1}{r}{60.2} & \multicolumn{1}{r}{60.25} & 0     &       & 1     & 44.7  & 44.69 & 0     &       & 1     & 163.7 & 163.67 & 0 \\
samp.frac & \multicolumn{1}{r}{3} & \multicolumn{1}{r}{90.9} & \multicolumn{1}{r}{30.30} & 0     &       & 3     & 67.0  & 22.33 & 0     &       & 3     & 245.2 & 81.74 & 0 \\
repl:samp.frac & \multicolumn{1}{r}{2} & \multicolumn{1}{r}{24.7} & \multicolumn{1}{r}{12.36} & 0     &       & 2     & 19.9  & 9.94  & 0     &       & 2     & 73.1  & 36.56 & 0 \\
Residuals               & \multicolumn{1}{r}{653} & \multicolumn{1}{r}{337.9} & \multicolumn{1}{r}{0.52} &       &       & 318   & 209.9 & 0.66  &       &       & 1332  & 829.3 & 0.62  &  \\
\bottomrule
\multicolumn{4}{l}{$^*$ values divided by $10^{12}$} &       &       &       &       &       &       &       &       &       &       &  \\
\end{tabular}%
 }
   \caption{Insurance data:  ANOVA tables comparing BCV 4x2, BCV 4x0 and RCV 16Rep with 10--CV SRS strategy.}
 \label{tab:ic_anova}%
\end{table}%
\renewcommand{\baselinestretch}{\bsln}

\subsection{Wine data}
The data set \citep{Wine} consists of 4898 variants of white wine, and the objective is to classify them into seven quality classes (measured on a Likert scale from three to nine) using 11 attributes. Additionally, we conducted regression analysis on the target variable as recommended by the authors. We present the results for both classification and regression tasks. The \ecv values were computed on a grid defined by the hyperparameter values outlined in Table \ref{tab:w_param}, using RFs with 1,500 trees.
%
\renewcommand{\baselinestretch}{0.9}
\begin{table}[H]
  \centering
{\scriptsize
\begin{tabular}{ll}
\toprule
Hyperparameter & Values \\
\midrule
mtry  & 3 6 9 \\
min.node.size & 3  5 10 20 \\
replace & T F \\
sample.fraction & 0.5 0.7 0.9 1.0 \\
\bottomrule
\end{tabular}%
 }
  \caption{Wine data: hyperparameter values used for tuning.}
  \label{tab:w_param}%
\end{table}%
\renewcommand{\baselinestretch}{\bsln}
Table \ref{tab:w_best2} shows that there is an almost perfect agreement between the top-performing settings obtained through BCV with two different designs of four blocks, BCV 2x2 and BCV 4x0, and RCV 16Rep, for both classification and regression predictions. The tuning favours sampling with replacement for classification and with for regression.

\renewcommand{\baselinestretch}{0.9}
\begin{table}[H]
  \centering
{\scriptsize
\begin{tabular}{rrrrrrl}
\toprule
\multicolumn{1}{l}{mtry} & \multicolumn{1}{l}{min.node.size} & \multicolumn{1}{l}{replace} & \multicolumn{1}{l}{sample.fraction} & \multicolumn{1}{l}{TreatNo} & \multicolumn{1}{l}{$\ecvb$} & Design \\
\midrule
\multicolumn{7}{c}{Regression} \\
3     & 3     & FALSE & 0.9   & 61    & 0.333 & \multirow{2}[1]{*}{BCV 2x2} \\
3     & 5     & FALSE & 0.9   & 64    & 0.338 &  \\
\midrule
3     & 3     & FALSE & 0.9   & 61    & 0.335 & \multirow{2}[2]{*}{BCV 4x0} \\
3     & 5     & FALSE & 0.9   & 64    & 0.340 &  \\
\midrule
3     & 3     & FALSE & 0.9   & 61    & 0.333 & \multirow{2}[2]{*}{RCV 8Rep} \\
3     & 5     & FALSE & 0.9   & 64    & 0.338 &  \\
\midrule
\multicolumn{7}{c}{Classification} \\
3     & 3     & TRUE  & 1     & 73    & 0.292 & \multirow{2}[1]{*}{BCV 2x2} \\
3     & 3     & TRUE  & 0.9   & 49    & 0.293 &  \\
\midrule
3     & 3     & TRUE  & 1     & 73    & 0.299 & \multirow{2}[2]{*}{BCV 4x0} \\
3     & 3     & TRUE  & 0.9   & 49    & 0.300 &  \\
\midrule
3     & 3     & TRUE  & 0.9   & 49    & 0.295 & \multirow{2}[2]{*}{RCV 8Rep} \\
3     & 3     & TRUE  & 1     & 73    & 0.296 &  \\
\bottomrule
\end{tabular}%
 }
  \caption{Wine data: the two settings that yielded the lowest $\ecvb$ for BCV 2x2, BCV 4x0 and RCV  16Rep, all using 10--folds and SRS strategy.}
  \label{tab:w_best2}%
\end{table}%
\renewcommand{\baselinestretch}{\bsln}
Table \ref{tab:w_anova} presents the ANOVA tables for BCV 4x2, BCV 8x0 and RCV Rep16 using 10-CV SRS. These highlight the comparable precision achieved by BCV with eight blocks in relation to RCV Rep16. The impact of RFseeds is minimal, while the influence of CVseeds is relatively modest compared to the effects of hyperparameters.
\renewcommand{\baselinestretch}{0.9}
\begin{table}[H]
  \centering
{\scriptsize
\begin{tabular}{lrrrrrrrrrrrrrr}
\toprule
      & \multicolumn{4}{c}{BCV 4x2}   &       & \multicolumn{4}{c}{BCV 8x0}   &       & \multicolumn{4}{c}{RCV 8Rep} \\
\cmidrule{2-5}\cmidrule{7-10}\cmidrule{12-15}source & Df    & $\text{SSE}^*$ & $\text{MSE}^*$ & Prob  &       & Df    & $\text{SSE}^*$ & $\text{MSE}^*$ & Prob  &       & Df    & $\text{SSE}^*$ & $\text{MSE}^*$ & Prob \\
\cmidrule{1-5}\cmidrule{7-10}\cmidrule{12-15}\multicolumn{15}{c}{Random Effects} \\
CVseeds                 & 3     & 2.88  & 0.960 & 0     &       & 7     & 3.46  & 0.495 & 0     &       &       &       &       &  \\
repl:CVseeds         & 3     & 0.01  & 0.002 & 0.97  &       & 7     & 0.01  & 0.001 & 1     &       &       &       &       &  \\
RFseeds                 & 1     & 0.00  & 0.001 & 0.73  &       &       &       &       &       &       &       &       &       &  \\
Total & 7     & 2.89  & 0.413 &       &       & 14    & 3.47  & 0.248 &       &       &       &       &       &  \\
\midrule
\multicolumn{15}{c}{Fixed Effects} \\
mtry                    & 2     & 1.58  & 0.789 & 0     &       & 2     & 1.17  & 0.587 & 0     &       & 2     & 0.79  & 0.394 & 0 \\
min.node & 3     & 152.56 & 50.853 & 0     &       & 3     & 148.17 & 49.390 & 0     &       & 3     & 246.34 & 82.113 & 0 \\
replace                 & 1     & 4.86  & 4.856 & 0     &       & 1     & 5.04  & 5.035 & 0     &       & 1     & 1.99  & 1.989 & 0 \\
samp.frac & 3     & 89.74 & 29.913 & 0     &       & 3     & 86.90 & 28.966 & 0     &       & 3     & 33.76 & 11.253 & 0 \\
repl:samp.frac & 2     & 0.19  & 0.096 & 0     &       & 2     & 0.18  & 0.088 & 0     &       & 2     & 4.32  & 2.162 & 0 \\
Residuals               & 653   & 8.491 & 0.013 &       &       & 646   & 7.747 & 0.012 &       &       & 660   & 21.626 & 0.033 &  \\
\bottomrule
\end{tabular}%
 }
  \caption{Wine data:  ANOVA tables for 10--CV SRS strategy comparing BCV 4x2, BCV 8x0 and RCV 8Rep for regression \ecv.}
  \label{tab:w_anova}%
\end{table}%
\renewcommand{\baselinestretch}{\bsln}

\begin{figure}[H]
  \centering
  \includegraphics[width=0.75\textwidth]{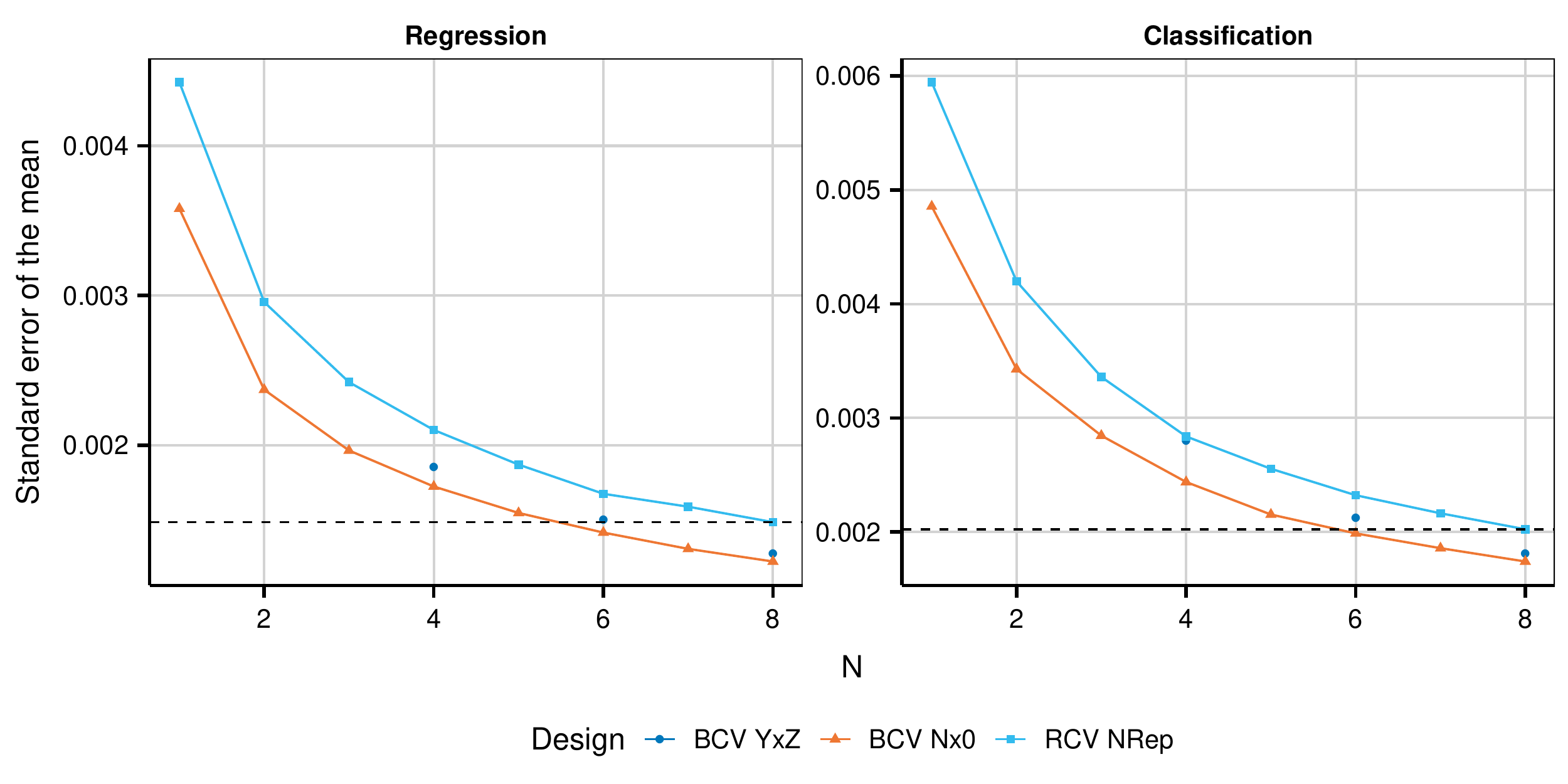}
  \caption{Wine data: comparison of std.err for classification and regression tasks using different 10--CV SRS variance reduction designs and number of runs. The horizontal lines mark the minimum std.err obtained with RCV.} \label{fig:w_stderr}%
\end{figure}

The std.errs shown in Figure \ref{fig:w_stderr} are rather small compared to the $\ecvb$ estimate shown in Table \ref{tab:w_best2}. BCV consistently outperforms RCV. However, due to the negligible effect of RFseed, BCV YxZ yields a slightly higher std.err than BCV Nx0 for the same number of blocks.

\subsection{Adult Data}
The task for this data set \citep{Adults} is to classify 30162 adults into low and high income using 13 socio--economic features. The \ecv values were computed on a grid defined by the hyperparameters values shown in Table \ref{tab:ad_param} using RFs computed  with 800 trees.
\renewcommand{\baselinestretch}{0.9}
\begin{table}[H]
  \centering
{\scriptsize
\begin{tabular}{ll}
\toprule
Hyperparameter & Values \\
\midrule
mtry  & 3  6 12 \\
min.node.size & 1  3  5 10 \\
replace & T F \\
sample.fraction & 0.5 0.7 0.9 1.0 \\
\bottomrule
\end{tabular}%
}
  \caption{Adult data: hyperparameter values used for tuning.}
\label{tab:ad_param}%
\end{table}%
\renewcommand{\baselinestretch}{\bsln}

Table \ref{tab:ad_best2} compares the two best settings obtained for tuning the RF using BCV 4x2, BCV 8x0 and RCV 16Rep, sampling 10 folds with SRS and STS. The \ecv values are very close to each other, but the optimal settings differ mainly by the value of the minimum.nod.size parameter.
\renewcommand{\baselinestretch}{0.9}
\begin{table}[H]
  \centering
{\scriptsize
\begin{tabular}{rrrrrrrrrrrrrrl}
\toprule
\multicolumn{6}{c}{10-Folds SRS}              &       & \multicolumn{6}{c}{10-Folds STS}              &       &  \\
\cmidrule{1-6}\cmidrule{8-13}\cmidrule{15-15}\multicolumn{1}{l}{mtry} & \multicolumn{1}{l}{min.nod} & \multicolumn{1}{l}{repl.} & \multicolumn{1}{l}{sam.frac} & \multicolumn{1}{l}{Sett.} & \multicolumn{1}{l}{$\ecvb$} &       & \multicolumn{1}{l}{mtry} & \multicolumn{1}{l}{min.nod} & \multicolumn{1}{l}{repl.} & \multicolumn{1}{l}{sam.frac} & \multicolumn{1}{l}{Sett.} & \multicolumn{1}{l}{$\ecvb$} &       & Design \\
\cmidrule{1-6}\cmidrule{8-13}\cmidrule{15-15}3     & 3     & F & 0.5   & 16    & 0.1356 &       & 3     & 5     & T  & 0.7   & 31    & 0.1357 &       & \multirow{2}[2]{*}{BCV Overall} \\
3     & 10    & F & 0.5   & 22    & 0.1356 &       & 3     & 10    & T  & 0.7   & 34    & 0.1357 &       &  \\
\cmidrule{1-6}\cmidrule{8-13}\cmidrule{15-15}3     & 3     & F & 0.5   & 16    & 0.1355 &       & 3     & 5     & T  & 0.7   & 31    & 0.1356 &       & \multirow{2}[2]{*}{BCv 4x2} \\
3     & 10    & T  & 0.7   & 34    & 0.1355 &       & 3     & 5     & T  & 0.5   & 7     & 0.1356 &       &  \\
\cmidrule{1-6}\cmidrule{8-13}\cmidrule{15-15}3     & 10    & F & 0.5   & 22    & 0.1356 &       & 3     & 10    & T  & 0.7   & 34    & 0.1355 &       & \multirow{2}[2]{*}{BCV 8x0} \\
3     & 10    & T  & 0.5   & 10    & 0.1356 &       & 3     & 5     & F & 0.5   & 19    & 0.1357 &       &  \\
\cmidrule{1-6}\cmidrule{8-13}\cmidrule{15-15}3     & 3     & F & 0.5   & 16    & 0.1353 &       & 3     & 3     & T  & 0.7   & 28    & 0.1356 &       & \multirow{2}[2]{*}{RCV 8Rep} \\
3     & 10    & T  & 0.7   & 34    & 0.1356 &       & 3     & 5     & T  & 0.5   & 7     & 0.1356 &       &  \\
\bottomrule
\end{tabular}%
 }
  \caption{Adult data: two best settings obtained with 10-CV and SRS and STS designs for different eight-rundesigns.}
  \label{tab:ad_best2}%
\end{table}%
\renewcommand{\baselinestretch}{\bsln}
Table \ref{tab:ad_anova} displays the ANOVA tables for $\ecvb$ obtained using 10--CV STS and the designs BCV 4x2, BCV 8x0, and RCV 8Rep. The blocked designs generally exhibit a lower MSE of the fixed effects. However, the magnitude of this improvement is relatively smaller compared to other cases, as the MSE of the blocks is small in comparison to that of the fixed effects.

\begin{figure}[H]
  \centering
  \includegraphics[width=0.75\textwidth]{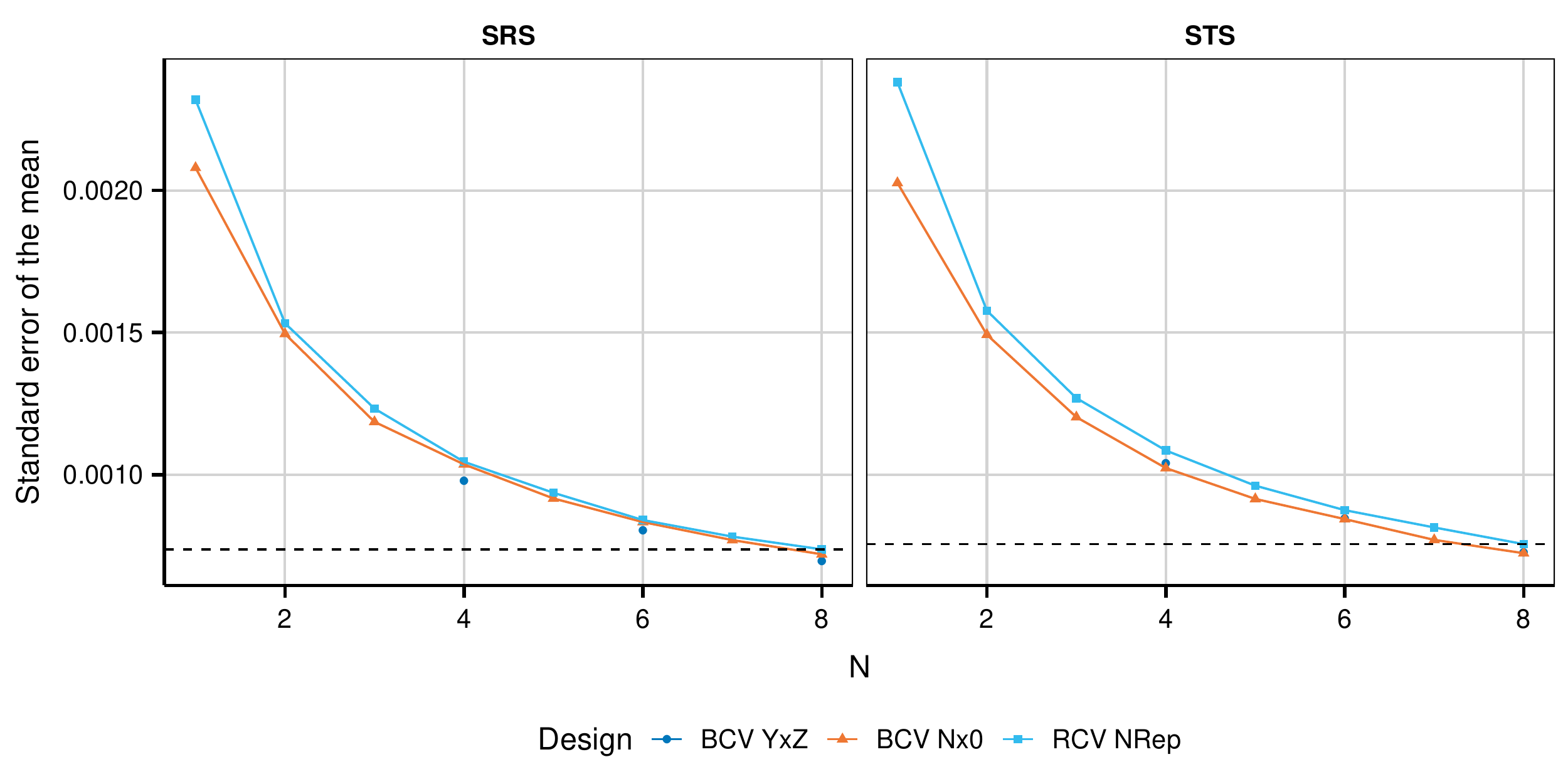}
  \caption{Adult data: comparison of the std.err for different designs and number of runs for 10--CV SRS and STS. The horizontal lines mark the minimum std.err obtained with RCV.} \label{fig:ad_stderr}%
\end{figure}

\renewcommand{\baselinestretch}{0.9}
\begin{table}[H]
  \centering
{\scriptsize
\begin{tabular}{llllllrrrrrrrrr}
\toprule
      & \multicolumn{4}{c}{BCV 4x2} &       & \multicolumn{4}{c}{BCV 8x0} &       & \multicolumn{4}{c}{RCV 8Rep} \\
\cmidrule{1-5}\cmidrule{7-10}\cmidrule{12-15}source & \multicolumn{1}{r}{Df} & \multicolumn{1}{r}{$\text{SSE}^*$} & \multicolumn{1}{r}{$\text{MSE}^*$} & \multicolumn{1}{r}{Prob} &       & Df    & $\text{SSE}^*$ & $\text{MSE}^*$ & Prob  &       & Df    & $\text{SSE}^*$ & $\text{MSE}^*$ & Prob \\
\cmidrule{1-5}\cmidrule{7-10}\cmidrule{12-15}\multicolumn{10}{c}{Random Effects}                                           &       &       &       &       &  \\
CVseeds & \multicolumn{1}{r}{3} & \multicolumn{1}{r}{0.6} & \multicolumn{1}{r}{0.2} & \multicolumn{1}{r}{0} &       & 7     & 6.5   & 0.9   & 0     &       &       &       &       &  \\
repl:CV & \multicolumn{1}{r}{3} & \multicolumn{1}{r}{0.6} & \multicolumn{1}{r}{0.2} & \multicolumn{1}{r}{0.97} &       & 7     & 0.2   & 0.0   & 1     &       &       &       &       &  \\
RFseeds & \multicolumn{1}{r}{1} & \multicolumn{1}{r}{0.0} & \multicolumn{1}{r}{0.0} & \multicolumn{1}{r}{0.73} &       &       &       &       &       &       &       &       &       &  \\
Total & \multicolumn{1}{r}{7} & \multicolumn{1}{r}{1.2} & \multicolumn{1}{r}{0.2} &       &       & 14    & 6.7   & 0.5   &       &       &       &       &       &  \\
\midrule
\multicolumn{15}{c}{Fixed Effects} \\
mtry                    & \multicolumn{1}{r}{2} & \multicolumn{1}{r}{1241.1} & \multicolumn{1}{r}{620.6} & \multicolumn{1}{r}{0} &       & 2     & 1131.8 & 565.9 & 0     &       & 2     & 1194.5 & 597.2 & 0 \\
min.node           & \multicolumn{1}{r}{3} & \multicolumn{1}{r}{55.3} & \multicolumn{1}{r}{18.4} & \multicolumn{1}{r}{0} &       & 3     & 51.6  & 17.2  & 0     &       & 3     & 51.3  & 17.1  & 0 \\
replace                 & \multicolumn{1}{r}{1} & \multicolumn{1}{r}{205.8} & \multicolumn{1}{r}{205.8} & \multicolumn{1}{r}{0} &       & 1     & 186.6 & 186.6 & 0     &       & 1     & 199.5 & 199.5 & 0 \\
samp.frac      & \multicolumn{1}{r}{3} & \multicolumn{1}{r}{311.5} & \multicolumn{1}{r}{103.8} & \multicolumn{1}{r}{0} &       & 3     & 277.2 & 92.4  & 0     &       & 3     & 305.2 & 101.7 & 0 \\
repl:samp.frac & \multicolumn{1}{r}{2} & \multicolumn{1}{r}{67.5} & \multicolumn{1}{r}{33.7} & \multicolumn{1}{r}{0} &       & 2     & 62.7  & 31.3  & 0     &       & 2     & 64.2  & 32.1  & 0 \\
Residuals               & \multicolumn{1}{r}{653} & \multicolumn{1}{r}{275.9} & \multicolumn{1}{r}{0.4} &       &       & 646   & 270.6 & 0.4   &       &       & 660   & 301.5 & 0.5   &  \\
\midrule
\multicolumn{6}{l}{$^*$ values multiplied by $10^5$} &       &       &       &       &       &       &       &       &  \\
\end{tabular}%
}
  \caption{Adult data: ANOVA tables for \ecv obtained with 10-fold STS and different eight--runs designs.}
  \label{tab:ad_anova}%
\end{table}%
\renewcommand{\baselinestretch}{\bsln}
Figure \ref{fig:ad_stderr} depicts the standard errors of $\ecvb$ obtained using the 10-fold SRS and STS strategies with the eight-block designs BCV YxZ, BCV Nx0, and RCV NRep. The standard errors are relatively small, and the improvement in the blocked designs is also modest. This can be attributed to the minimal impact of blocking as mentioned earlier.
\section{Concluding remarks}\label{sec:remarks}

Our research clearly shows that integrating blocking into grid search significantly reduces the standard error of the estimated $\ecvb$, outperforming the results achieved through simple RCV. Blocking not only enhances precision but also provides more accurate estimates of hyperparameters' effects. We have provided a solid theoretical rationale for this variance reduction and supported our assertions with convincing examples where BCV outperformed RCV.

Although we demonstrated the effectiveness of BCV on tuning RFs, BCV can be employed to tune any type of learner, thereby extending its applicability across the entire field of hyperparameter tuning.

Through our extensive experiments, we consistently observed the efficacy of blocking with respect to CV partitioning when tuning RFs. However, it is noteworthy that the magnitude of improvement over simple randomization diminishes as the sample size increases. Blocking based on random behavior is rarely necessary. Furthermore, we have found that, in some cases, the interaction between CV partitioning and the choice of subsampling (with or without replacement) has a significant impact on reducing variance.

BCV has several advantages over RCV:
\begin{itemize}
  \item It requires fewer runs to achieve the same standard error of estimated \ecv for different settings;
  \item it provides more precise estimates of hyperparameters' effects;
  \item it is computationally more efficient as it requires fewer data partitions and can be more efficiently parallelized;
  \item it can be used to compare various outcomes of the learners, such as full residuals, leveraging the advantages of reduced variability due to blocking.
\end{itemize}

Given the compelling array of advantages outlined above, we strongly believe that BCV should be the preferred choice over RCV in all relevant analyses. Essentially, BCV improves the estimates over RCV with reduced computational effort. Opting for BCV is a sensible decision that maximizes the benefits obtained from the analysis, leading to robust and reliable results.

In this paper, our main objective was to demonstrate the effectiveness of BCV in improving the precision of \ecv estimation. However, the usefulness of BCV extends beyond simple grid search for hyperparameter tuning, making it applicable to a wider range of tasks.

In our manuscript, we intentionally did not delve into the analysis of \ecv values obtained with the grid. However, it is  worth mentioning that BCV offers valuable opportunities for post-hoc analyses, such as verifying the significance of subsets of hyperparameters. By incorporating blocking, the analyses can still benefit from the enhanced precision achieved through this technique.

Moreover, the advantages of BCV are useful also in sensitivity analysis. While in our current article we treated hyperparameter values as nominal levels to accommodate nonlinear behaviors without delving into the complexities of sensitivity analysis, it is important to note that when employing \ecv for fine-tuning a learner, blocking still enhances the precision of the analysis. An intriguing possibility is to implement blocking with respect to CV partitioning, using second-order rotatable designs proposed in \cite{mor}.

BCV also has an important application in model selection. It enables a fair comparison among different predictive learners by adhering to the fundamental principle of "comparing like with like" within the blocking framework, as outlined in Equation (4) of \cite{ber}.
\bibliographystyle{plainnat}

\renewcommand\refname{Data Sources}

\end{document}


%% file: Tuning_MACROS.TeX
\newcommand{\ecv}{{$Err^{CV}$}}
\newcommand{\ecvm}{{Err^{CV}}}

\newcommand{\ecvb}{{\protect\widebar{Err}^{CV}}}

\newcommand{\ercv}{{${Err}^{RCV}$}}
\newcommand{\ebcvs}[1]{{Err}^{BCV}_{#1}}
\newcommand{\ercvs}[1]{{Err}^{RCV}_{#1}}

\newcommand{\ebcvbs}[1]{\widebar{Err}^{BCV}_{#1}}
\newcommand{\ercvbs}[1]{\widebar{Err}^{RCV}_{#1}}

\newcommand{\bth}{{\boldsymbol{\theta}}}
\newcommand{\vare}{{\varepsilon}}

\makeatletter
\let\save@mathaccent\mathaccent
\newcommand*\if@single[3]{%
  \setbox0\hbox{${\mathaccent"0362{#1}}^H$}%
  \setbox2\hbox{${\mathaccent"0362{\kern0pt#1}}^H$}%
  \ifdim\ht0=\ht2 #3\else #2\fi
  }
\newcommand*\rel@kern[1]{\kern#1\dimexpr\macc@kerna}
\newcommand*\widebar[1]{\@ifnextchar^{{\wide@bar{#1}{0}}}{\wide@bar{#1}{1}}}
\newcommand*\wide@bar[2]{\if@single{#1}{\wide@bar@{#1}{#2}{1}}{\wide@bar@{#1}{#2}{2}}}
\newcommand*\wide@bar@[3]{%
  \begingroup
  \def\mathaccent##1##2{%
    \let\mathaccent\save@mathaccent
    \if#32 \let\macc@nucleus\first@char \fi
    \setbox\z@\hbox{$\macc@style{\macc@nucleus}_{}$}%
    \setbox\tw@\hbox{$\macc@style{\macc@nucleus}{}_{}$}%
    \dimen@\wd\tw@
    \advance\dimen@-\wd\z@
    \divide\dimen@ 3
    \@tempdima\wd\tw@
    \advance\@tempdima-\scriptspace
    \divide\@tempdima 10
    \advance\dimen@-\@tempdima
    \ifdim\dimen@>\z@ \dimen@0pt\fi
    \rel@kern{0.6}\kern-\dimen@
    \if#31
      \overline{\rel@kern{-0.6}\kern\dimen@\macc@nucleus\rel@kern{0.4}\kern\dimen@}%
      \advance\dimen@0.4\dimexpr\macc@kerna
      \let\final@kern#2%
      \ifdim\dimen@<\z@ \let\final@kern1\fi
      \if\final@kern1 \kern-\dimen@\fi
    \else
      \overline{\rel@kern{-0.6}\kern\dimen@#1}%
    \fi
  }%
  \macc@depth\@ne
  \let\math@bgroup\@empty \let\math@egroup\macc@set@skewchar
  \mathsurround\z@ \frozen@everymath{\mathgroup\macc@group\relax}%
  \macc@set@skewchar\relax
  \let\mathaccentV\macc@nested@a
  \if#31
    \macc@nested@a\relax111{#1}%
  \else
    \def\gobble@till@marker##1\endmarker{}%
    \futurelet\first@char\gobble@till@marker#1\endmarker
    \ifcat\noexpand\first@char A\else
      \def\first@char{}%
    \fi
    \macc@nested@a\relax111{\first@char}%
  \fi
  \endgroup
}
\makeatother

%% file: Tuning_paper_v5_ML.bbl
\begin{thebibliography}{38}
\providecommand{\natexlab}[1]{#1}
\providecommand{\url}[1]{\texttt{#1}}
\expandafter\ifx\csname urlstyle\endcsname\relax
  \providecommand{\doi}[1]{doi: #1}\else
  \providecommand{\doi}{doi: \begingroup \urlstyle{rm}\Url}\fi

\bibitem[Allen(1974)]{all}
David~M Allen.
\newblock The relationship between variable selection and data agumentation and
  a method for prediction.
\newblock \emph{Technometrics}, 16\penalty0 (1):\penalty0 125--127, 1974.

\bibitem[Arlot and Celisse(2010)]{arl}
Sylvain Arlot and Alain Celisse.
\newblock A survey of cross-validation procedures for model selection.
\newblock \emph{Statistics Surveys}, 4:\penalty0 40--7, 2010.

\bibitem[Bates et~al.(2021)Bates, Hastie, and Tibshirani]{bat}
Stephen Bates, Trevor Hastie, and Robert Tibshirani.
\newblock Cross-validation: What does it estimate and how well does it do it?
\newblock \emph{arXiv preprint arXiv:2104.00673}, 2021.

\bibitem[Bengio and Grandvalet(2004)]{ben}
Yoshua Bengio and Yves Grandvalet.
\newblock No unbiased estimator of the variance of k-fold cross-validation.
\newblock \emph{Journal of Machine Learning Research}, 5, Sep. 2004.

\bibitem[Bergstra and Bengio(2012)]{ber}
James Bergstra and Yoshua Bengio.
\newblock Random search for hyper-parameter optimization.
\newblock \emph{Journal of machine learning research}, 13\penalty0 (2), 2012.

\bibitem[Boulesteix and Strobl(2009)]{bou}
Anne-Laure Boulesteix and Carolin Strobl.
\newblock Optimal classifier selection and negative bias in error rate
  estimation: an empirical study on high-dimensional prediction.
\newblock \emph{BMC medical research methodology}, 9\penalty0 (1):\penalty0
  1--14, 2009.

\bibitem[Box et~al.(2005)Box, Hunter, and Hunter]{box}
George~E.P. Box, J~Stuart Hunter, and William~G Hunter.
\newblock \emph{Statistics for experimenters: Statistics For Experimenters:
  Design, Innovation, and discovery}.
\newblock Wiley series in probability and statistics. Wiley Hoboken, NJ, 2nd
  edition, 2005.

\bibitem[Breiman et~al.(1984)Breiman, Friedman, Olshen, and Stone]{bre84}
Leo Breiman, Jerome~H. Friedman, Richard~A. Olshen, and Charles~J. Stone.
\newblock \emph{Classification and Regression Trees}.
\newblock Routledge, 1984.

\bibitem[Burman(1989)]{bur}
Prabir Burman.
\newblock A comparative study of ordinary cross-validation, v-fold
  cross-validation and the repeated learning-testing methods.
\newblock \emph{Biometrika}, 76\penalty0 (3):\penalty0 503--514, 1989.

\bibitem[Cortez et~al.(2009)Cortez, Cerdeira, Almeida, Matos, and Reis]{Wine}
Paulo Cortez, Ant{\'o}nio Cerdeira, Fernando Almeida, Telma Matos, and Joaquim
  Reis.
\newblock Modeling wine preferences by data mining from physicochemical
  properties.
\newblock \emph{Decision Support Systems}, 47\penalty0 (4):\penalty0 547--553,
  2009.
\newblock data \url{https://archive.ics.uci.edu/ml/datasets/wine+quality}.

\bibitem[Cox and Reid(2000)]{cox}
D.R. Cox and N.~Reid.
\newblock \emph{{The Theory of the Design of Experiments}}.
\newblock Chapman \& Hall/CRC, 2000.

\bibitem[Dietterich(1998)]{die}
Thomas~G. Dietterich.
\newblock Approximate statistical tests for comparing supervised classification
  learning algorithms.
\newblock \emph{Neural Computation}, 10\penalty0 (7):\penalty0 1895–1923, Oct
  1998.

\bibitem[Efron(1983)]{efr83}
Bradley Efron.
\newblock Estimating the error rate of a prediction rule: improvement on
  cross-validation.
\newblock \emph{Journal of the American statistical association}, 78\penalty0
  (382):\penalty0 316--331, 1983.

\bibitem[Fisher(1937)]{fis}
Ronald~A Fisher.
\newblock \emph{The design of experiments}.
\newblock Oliver \& Boyd, 1937.

\bibitem[Frossard and Renaud(2021)]{fro}
Jaromil Frossard and Olivier Renaud.
\newblock Permutation tests for regression, anova, and comparison of signals:
  The permuco package.
\newblock \emph{Journal of Statistical Software}, 99\penalty0 (15):\penalty0
  1–32, 2021.

\bibitem[Good(2013)]{good}
Phillip Good.
\newblock \emph{Permutation tests: a practical guide to resampling methods for
  testing hypotheses}.
\newblock Springer Science \& Business Media, 2013.

\bibitem[Gorman et~al.(2014)Gorman, Williams, and Fraser]{Penguin}
KB~Gorman, TD~Williams, and WR~Fraser.
\newblock Ecological sexual dimorphism and environmental variability within a
  community of antarctic penguins (genus pygoscelis).
\newblock \emph{PLoS ONE}, 9\penalty0 (3), 2014.
\newblock data
  \url{https://www.kaggle.com/datasets/parulpandey/palmer-archipelago-antarctica-penguin-data}.

\bibitem[Gorman and Sejnowski(1988)]{Sonar}
Russell~P Gorman and Terrence~J Sejnowski.
\newblock Analysis of hidden units in a layered network trained to classify
  sonar targets.
\newblock \emph{Neural Networks}, 1:\penalty0 75--89, 1988.
\newblock data
  \url{http://archive.ics.uci.edu/ml/datasets/connectionist+bench+(sonar,+mines+vs.+rocks)}.

\bibitem[Harrison and Rubinfeld(1978)]{Boston}
David Harrison and Daniel~L Rubinfeld.
\newblock Hedonic prices and the demand for clean air.
\newblock \emph{Journal of Environmental Economics and Management}, 5:\penalty0
  81--102, 1978.
\newblock data
  \url{https://www.kaggle.com/datasets/vikrishnan/boston-house-prices/versions/1?resource=download}.

\bibitem[Hastie et~al.(2017)Hastie, Tibshirani, Friedman, and Friedman]{esl}
Trevor Hastie, Robert Tibshirani, Jerome Friedman, and J.~H. Friedman.
\newblock \emph{The Elements of Statistical Learning: Data Mining, Inference,
  and Prediction}.
\newblock Springer Series in Statistics Ser., Jun 2017.

\bibitem[Kennedy(1995)]{ken}
Fetter~E Kennedy.
\newblock Randomization tests in econometrics.
\newblock \emph{Journal of Business \& Economic Statistics}, 13\penalty0
  (1):\penalty0 85--94, 1995.

\bibitem[Kim(2009)]{kim}
Ji-Hyun Kim.
\newblock Estimating classification error rate: Repeated cross-validation,
  repeated hold-out and bootstrap.
\newblock \emph{Computational statistics \& data analysis}, 53\penalty0
  (11):\penalty0 3735--3745, 2009.

\bibitem[Kohavi(1996)]{Adults}
Ron Kohavi.
\newblock Scaling up the accuracy of naive-bayes classifiers: A decision-tree
  hybrid.
\newblock In \emph{Proceedings of the Second International Conference on
  Knowledge Discovery and Data Mining}, KDD'96, page 202â€“207, Portland,
  Oregon, 1996. AAAI Press.
\newblock data \url{https://archive.ics.uci.edu/ml/datasets/Adult}.

\bibitem[Kohavi et~al.(1995)]{koh}
Ron Kohavi et~al.
\newblock A study of cross-validation and bootstrap for accuracy estimation and
  model selection.
\newblock In \emph{International Joint Conference on Artificial Intelligence},
  volume 14:2, pages 1137--1145. Montreal, Canada, 1995.

\bibitem[Kuhn(2008)]{kuhcar}
Max Kuhn.
\newblock Building predictive models in r using the caret package.
\newblock \emph{Journal of Statistical Software}, 28\penalty0 (5):\penalty0
  1–26, 2008.

\bibitem[Kuhn and Johnson(2013)]{kuh}
Max Kuhn and Kjell Johnson.
\newblock \emph{Applied predictive modeling}, volume~26.
\newblock Springer, New York, 2013.

\bibitem[Lantz(2019)]{Insurance}
B.~Lantz.
\newblock \emph{Machine Learning with R: Expert techniques for predictive
  modeling, 3rd Edition}.
\newblock Packt Publishing, 2019.
\newblock data \url{https://www.kaggle.com/datasets/mirichoi0218/insurance}.

\bibitem[Lujan-Moreno et~al.(2018)Lujan-Moreno, Howard, Rojas, and
  Montgomery]{mor}
Gustavo~A Lujan-Moreno, Phillip~R Howard, Omar~G Rojas, and Douglas~C
  Montgomery.
\newblock Design of experiments and response surface methodology to tune
  machine learning hyperparameters, with a random forest case-study.
\newblock \emph{Expert Systems with Applications}, 109:\penalty0 195--205,
  2018.

\bibitem[Molinaro et~al.(2005)Molinaro, Simon, and Pfeiffer]{mol}
Annette~M. Molinaro, Richard Simon, and Ruth~M. Pfeiffer.
\newblock Prediction error estimation: a comparison of resampling methods.
\newblock \emph{Bioinformatics}, 21\penalty0 (15):\penalty0 3301--3307, 2005.

\bibitem[Montgomery(2008)]{mon}
Douglas~C. Montgomery.
\newblock \emph{{Design and Analysis of Experiments}}.
\newblock Wiley, 7th edition, 2008.

\bibitem[{R Core Team}(2020)]{R}
{R Core Team}.
\newblock \emph{R: A Language and Environment for Statistical Computing}.
\newblock R Foundation for Statistical Computing, Vienna, Austria, 2020.
\newblock URL \url{https://www.R-project.org/}.

\bibitem[Stone(1974)]{sto74}
Mervyn Stone.
\newblock Cross-validatory choice and assessment of statistical predictions.
\newblock \emph{Journal of the royal statistical society: Series B
  (Methodological)}, 36\penalty0 (2):\penalty0 111--133, 1974.

\bibitem[Vanwinckelen and Blockeel(2012)]{van}
Gitte Vanwinckelen and Hendrik Blockeel.
\newblock On estimating model accuracy with repeated cross-validation.
\newblock In \emph{BeneLearn 2012: Proceedings of the 21st Belgian-Dutch
  conference on machine learning}, pages 39--44, 2012.

\bibitem[Varma and Simon(2006)]{var}
Sudhir Varma and Richard Simon.
\newblock Bias in error estimation when using cross-validation for model
  selection.
\newblock \emph{BMC bioinformatics}, 7\penalty0 (1):\penalty0 1--8, 2006.

\bibitem[Wheeler and Torchiano(2016)]{lmPerm}
Bob Wheeler and Marco Torchiano.
\newblock \emph{lmPerm: Permutation Tests for Linear Models}, 2016.
\newblock R package version 2.1.0.

\bibitem[Wright and Ziegler(2017)]{ranger}
Marvin~N. Wright and Andreas Ziegler.
\newblock {ranger}: A fast implementation of random forests for high
  dimensional data in {C++} and {R}.
\newblock \emph{Journal of Statistical Software}, 77\penalty0 (1):\penalty0
  1--17, 2017.
\newblock \doi{10.18637/jss.v077.i01}.

\bibitem[Wu and Hamada(2011)]{wu}
CF~Jeff Wu and Michael~S Hamada.
\newblock \emph{Experiments: planning, analysis, and optimization}.
\newblock John Wiley \& Sons, 3rd edition, 2011.

\bibitem[Yousef(2021)]{you21}
Waleed~A Yousef.
\newblock Estimating the standard error of cross-validation-based estimators of
  classifier performance.
\newblock \emph{Pattern Recognition Letters}, 146:\penalty0 115--125, 2021.

\end{thebibliography}

\begin{thebibliography}{6}
\providecommand{\natexlab}[1]{#1}
\providecommand{\url}[1]{\texttt{#1}}
\expandafter\ifx\csname urlstyle\endcsname\relax
  \providecommand{\doi}[1]{doi: #1}\else
  \providecommand{\doi}{doi: \begingroup \urlstyle{rm}\Url}\fi

\bibitem[Cortez et~al.(2009)Cortez, Cerdeira, Almeida, Matos, and Reis]{Wine}
Paulo Cortez, Ant{\'o}nio Cerdeira, Fernando Almeida, Telma Matos, and Joaquim
  Reis.
\newblock Modeling wine preferences by data mining from physicochemical
  properties.
\newblock \emph{Decision Support Systems}, 47\penalty0 (4):\penalty0 547--553,
  2009.
\newblock data \url{https://archive.ics.uci.edu/ml/datasets/wine+quality}.

\bibitem[Gorman et~al.(2014)Gorman, Williams, and Fraser]{Penguin}
KB~Gorman, TD~Williams, and WR~Fraser.
\newblock Ecological sexual dimorphism and environmental variability within a
  community of antarctic penguins (genus pygoscelis).
\newblock \emph{PLoS ONE}, 9\penalty0 (3), 2014.
\newblock data
  \url{https://www.kaggle.com/datasets/parulpandey/palmer-archipelago-antarctica-penguin-data}.

\bibitem[Gorman and Sejnowski(1988)]{Sonar}
Russell~P Gorman and Terrence~J Sejnowski.
\newblock Analysis of hidden units in a layered network trained to classify
  sonar targets.
\newblock \emph{Neural Networks}, 1:\penalty0 75--89, 1988.
\newblock data
  \url{http://archive.ics.uci.edu/ml/datasets/connectionist+bench+(sonar,+mines+vs.+rocks)}.

\bibitem[Harrison and Rubinfeld(1978)]{Boston}
David Harrison and Daniel~L Rubinfeld.
\newblock Hedonic prices and the demand for clean air.
\newblock \emph{Journal of Environmental Economics and Management}, 5:\penalty0
  81--102, 1978.
\newblock data
  \url{https://www.kaggle.com/datasets/vikrishnan/boston-house-prices/versions/1?resource=download}.

\bibitem[Kohavi(1996)]{Adults}
Ron Kohavi.
\newblock Scaling up the accuracy of naive-bayes classifiers: A decision-tree
  hybrid.
\newblock In \emph{Proceedings of the Second International Conference on
  Knowledge Discovery and Data Mining}, KDD'96, page 202--207, Portland,
  Oregon, 1996. AAAI Press.
\newblock data \url{https://archive.ics.uci.edu/ml/datasets/Adult}.

\bibitem[Lantz(2019)]{Insurance}
Brett Lantz.
\newblock \emph{Machine Learning with R: Expert techniques for predictive
  modeling, 3rd Edition}.
\newblock Packt Publishing, 2019.
\newblock data \url{https://www.kaggle.com/datasets/mirichoi0218/insurance}.


\end{thebibliography}
